\documentclass[journal,twocolumn]{IEEEtran}
\usepackage[usenames]{color}

\definecolor{plum}  {rgb}{.4,0,.4}
\definecolor{BrickRed} {rgb}{0.6,0,0}

\usepackage[plainpages=false,pdfpagelabels,colorlinks=true,linkcolor=BrickRed,citecolor=plum]{hyperref}

\usepackage{amsmath,amssymb,epsfig,amsthm}

\usepackage{verbatim}
\usepackage{algorithm,algorithmic}
\usepackage{graphicx}

\newtheorem{theorem}{Theorem}
\newtheorem{lemma}{Lemma}

\newtheorem{remark}{Remark}

\newtheorem{example}{Example}
\newtheorem{definition}{Definition}

\newtheorem{assumption}{Assumption}

\def\argmin{\mathop{\arg \min}}

\def\diam{\operatorname{diam}}

\def\cB{{\cal B}}
\def\cC{{\cal C}}
\def\cD{{\cal D}}
\def\cF{{\cal F}}
\def\cG{{\cal G}}
\def\cP{{\cal P}}
\def\cR{{\cal R}}
\def\cU{{\mathsf U}}
\def\cS{{\mathsf S}}
\def\cX{{\mathsf X}}

\def\cZ{{\mathsf Z}}

\def\sp{{\pi}}

\def\su{{\mu}}

\def\reals{{\mathbb R}}

\def\expect{{\mathbb E}}

\def\ave#1{\langle #1 \rangle}
\def\Ave#1{\Big\langle #1 \Big\rangle}

\def\bd#1{\boldsymbol{#1}}
\def\wh#1{\widehat{#1}}
\def\td#1{\widetilde{#1}}

\def\btheta{\bd{\theta}}
\def\bhtheta{\wh{\btheta}}

\def\Int{\operatorname{Int}}
\def\sgn{\operatorname{sgn}}

\def\deq{\triangleq}

\def\lf{\wh{\ell}}
\def\Lf{\wh{L}}
\def\Df{\Delta}

\def\FHTAGN{{\sc fhtagn}}

\begin{document}

\title{Sequential Anomaly Detection in the Presence of Noise and Limited Feedback}

\author{Maxim Raginsky,~\IEEEmembership{Member,~IEEE}, Rebecca
  Willett,~\IEEEmembership{Senior Member,~IEEE}, Corinne Horn, Jorge
  Silva,~\IEEEmembership{Member,~IEEE}, and Roummel
  Marcia,~\IEEEmembership{Member,~IEEE} \thanks{This work was
    supported by NSF CAREER Award No.\ CCF-06-43947, NSF Award No.\
    DMS-08-11062, DARPA Grant No.\ HR0011-07-1-003, and ARO Grant
    W911NF-09-1-0262. Portions of this work were presented at the IEEE
    International Symposium on Information Theory, Seoul, Korea,
    June/July 2009.}  \thanks{M. Raginsky was with the Department of
    Electrical and Computer Engineering, Duke University, Durham, NC
    27708, USA. He is now with the Department of Electrical and
    Computer Engineering and the Coordinated Science Laboratory,
    University of Illinois at Urbana-Champaign, Urbana, IL 6 801 USA
    (e-mail: maxim@illinois.edu).}  \thanks{R. Willett and J. Silva
    are with the Department of Electrical and Computer Engineering,
    Duke University, Durham, NC 27708 USA (e-mail:
    m.raginsky@duke.edu, willett@duke.edu, jg.silva@duke.edu).}
  \thanks{C. Horn was with the Department of Electrical and Computer
    Engineering, Duke University, Durham, NC 27708 USA. She is now
    with the Department of Electrical and Computer Engineering,
    Stanford University, Stanford, CA 94305 USA (e-mail:
    cehorn@stanford.edu).}  \thanks{R. Marcia is with the School of
    Natural Sciences, University of California, Merced, CA 95343 USA
    (e-mail: rmarcia@ucmerced.edu).}}

\maketitle

\begin{abstract}%
  This paper describes a methodology for detecting anomalies from
  sequentially observed and potentially noisy data. The proposed
  approach consists of two main elements: (1) {\em filtering}, or
  assigning a belief or likelihood to each successive measurement
  based upon our ability to predict it from previous noisy
  observations, and (2) {\em hedging}, or flagging potential anomalies
  by comparing the current belief against a time-varying and
  data-adaptive threshold. The threshold is adjusted based on the
  available feedback from an end user. Our algorithms, which combine
  universal prediction with recent work on online convex programming,
  do not require computing posterior distributions given all current
  observations and involve simple primal-dual parameter updates. At
  the heart of the proposed approach lie exponential-family models
  which can be used in a wide variety of contexts and applications,
  and which yield methods that achieve sublinear per-round regret
  against both static and slowly varying product distributions with
  marginals drawn from the same exponential family. Moreover, the
  regret against static distributions coincides with the minimax value
  of the corresponding online strongly convex game. We also prove
  bounds on the number of mistakes made during the hedging step
  relative to the best offline choice of the threshold with access to
  all estimated beliefs and feedback signals. We validate the theory
  on synthetic data drawn from a time-varying distribution over binary
  vectors of high dimensionality, as well as on the Enron email
  dataset.
 \\

\end{abstract}

\thispagestyle{empty}

\section{Introduction}
\label{sec:intro}

\PARstart{I}{n this} paper, we explore the performance of online anomaly detection
methods built on sequential probability assignment and dynamic
thresholding based on limited feedback. We assume we sequentially
monitor the state of some system of interest. At each time step, we
observe a possibly {\em noise-corrupted} version $z_t$ of the current state $x_t$, and need to infer whether $x_t$ is {\em anomalous} relative to the
actual sequence $x^{t-1} = (x_1,\ldots,x_{t-1})$ of the past states. This inference is encapsulated
in a binary decision $\wh{y}_t$, which can be either $-1$ (non-anomalous or nominal behavior) or
$+1$ (anomalous behavior). After announcing our decision, we may
occasionally receive {\em feedback} on the ``true'' state of affairs
and use it to adjust the future behavior of the decision-making
mechanism.

Our inference engine should make good use of this
feedback, whenever it is available, to improve its future
performance. One reasonable way to do it is as follows. Having
observed $z^{t-1}$ (but not $z_t$), we can use this observation to
assign ``beliefs" or ``likelihoods" to the clean state $x_t$. Let us
denote this likelihood assignment as $p_t(x_t | z^{t-1})$. Then, if we
actually had access to the clean observation $x_t$, we could evaluate
$p_t = p_t(x_t|z^{t-1})$ and declare an anomaly ($\wh{y}_t = +1$) if
$p_t < \tau_t$, where $\tau_t$ is some positive threshold; otherwise
we would set $\wh{y}_t = -1$ (no anomaly at time $t$). This approach
is based on the intuitive idea that a new observation $x_t$ should be
declared anomalous if it is very unlikely based on our past knowledge
(namely, $z^{t-1}$). In other words, observations are considered
anomalous if they are in a portion of the observation domain which has
very low likelihood according to the best probability model that can
be assigned to them on the basis of previously seen observations. (In
fact, anomaly detection algorithms based on density level sets revolve
around precisely this kind of reasoning.) The complication here,
however, is that we do not actually observe $x_t$, but rather its
noise-corrupted version $z_t$. Thus, we settle instead for an {\em
  estimate} $\wh{p}_t$ of $p_t$ based on $z_t$ and compare this
estimate against $\tau_t$. If we receive feedback $y_t$ at time $t$
and it differs from our label $\wh{y}_t$, then we adjust the threshold
appropriately.

\subsection{Contributions}

There are several challenging aspects inherent in the problem of
sequential anomaly detection:
\begin{itemize}
\item The observations cannot be assumed to be independent,
  identically distributed, or even come from a realization of a stochastic
  process. In particular, an adversary may be injecting false
  data into the sequence of observations to cripple our anomaly
  detection system.
\item Observations may be contaminated by noise or be observed through an imperfect
  communication channel.
\item Declaring observations anomalous if their likelihoods fall below some threshold is a popular and effective strategy for anomaly detection, but setting this threshold is a notoriously difficult problem.
\item Obtaining feedback on the quality of automated anomaly detection is costly as it generally involves considerable effort by a human expert or analyst. Thus, if we have an option to request such feedback at any time step, we should exercise this option sparingly and keep the number of requests to a minimum. Alternatively, the times when we receive feedback may be completely arbitrary and not under our control at all --- for instance, we may receive feedback only when we declare false positives or miss true anomalies.
\end{itemize}
In this paper, we propose a general methodology for addressing these challenges.  With apologies to H.P.~Lovecraft \cite{fhtagn}, we will call our proposed framework \FHTAGN, or \underline{F}iltering and \underline{H}edging for \underline{T}ime-varying \underline{A}nomaly reco\underline{GN}ition. More specifically, the two components that make up \FHTAGN\ are:
\begin{itemize}
	\item {\em Filtering} --- the sequential process of updating {\em beliefs} on the next state of the system based on the noisy observed past. The term ``filtering'' comes from statistical signal processing \cite{BaiCri09} and is intended to signify the fact that the beliefs of interest concern the unobservable {\em actual} system state, yet can only be computed in a {\em causal manner} from its noise-corrupted observations.
	
	\item {\em Hedging} --- the sequential process of flagging potential anomalies by comparing the current belief against a time-varying threshold. The rationale for this approach comes from the intuition that a behavior we could not have predicted well based on the past is likely to be anomalous. The term ``hedging'' is meant to indicate the fact that the threshold is dynamically raised or lowered, depending on the type of the most recent mistake (a false positive or a missed anomaly) made by our inference engine.
\end{itemize}
Rather than explicitly modeling the evolution of the system state and
then designing methods for that model (e.g.,~using Bayesian updates
\cite{BaiCri09,Sha09}), we adopt an ``individual sequence'' (or
``universal prediction" \cite{MerFed98}) perspective and strive to
perform provably well on any individual observation sequence in the
sense that our per-round performance approaches that of the best {\em
  offline} method with access to the entire data sequence.  This
approach allows us to sidestep challenging statistical issues
associated with dependent observations or dynamic and evolving
probability distributions, and is robust to noisy observations. 
We
make the following contributions:

{\em 1)}
We  cast both
  filtering and hedging as instances of {\em Online Convex
    Programming} (or OCP), as defined by Zinkevich \cite{Zin03}. This
  will permit us to implement both of these ingredients of \FHTAGN\
  using a powerful primal-dual method of Mirror Descent
  \cite{NemYud83,BecTeb03} and quantify their performance in a unified
  manner via regret bounds relative to the best offline strategy with
  access to the full observation sequence.

  {\em 2)} We show that the filtering step can be implemented as a
  sequential assignment of beliefs, or probabilities, to the system
  state based on the past noisy observations, where the probabilities
  are computed according to an exponential family model whose natural
  parameter is dynamically determined based on the past. We present a
  strategy based on mirror descent for sequentially assigning a
  time-varying product distribution with exponential-family marginals
  to the observed noisy sequence of system states and prove regret
  bounds relative to the best i.i.d.\ model as well as to the best
  sufficiently slowly changing model that can be assigned to the
  observation sequence in hindsight. These regret bounds improve and
  extend our preliminary results \cite{raginsky_OCP_isit09}; in
  addition to tightening the bounds presented in that work, we extend
  the results to more general settings in which data may be corrupted
  by noise. The main thing to keep in mind about the
  individual-sequence setting is that neither the sequence of
  probability assignments nor the best model that can be chosen
  offline should be interpreted as estimates of some ``true"
  stochastic process model of the observation sequence. Rather, both
  should be viewed as {\em algorithmic strategies} for predicting the
  next observation given the past (cf.\ the survey paper by Merhav and
  Feder \cite{MerFed98} for more details on the differences and the
  similarities between the more familiar probabilistic setting and the
  deterministic, individual-sequence setting used in this paper).
     
{\em 3)}
 We show that the hedging step can be implemented as a sequential
  selection of the critical threshold, such that whenever the
  estimated belief for the current state falls below this threshold,
  we declare an anomaly. We develop methods to incorporate available feedback and establish
  regret-type bounds on the number of mistakes relative to the best
  threshold that can be selected in hindsight with access to the
  entire sequence of assigned beliefs and feedback.

As described in the useful survey by Chandola et al.~\cite{anomalySurvey},
several methods for anomaly detection have been developed using
supervised \cite{SteHusSco05}, semi-supervised
\cite{ScoBla09}, and unsupervised
\cite{densityLevelSets} learning methods.  In the online,
individual-sequence setting we adopt, however, there is no {\em
  intrinsic} notion of what constitutes an anomaly.  Instead, we focus
on {\em extrinsic} anomalous behavior relative to the best {\em model}
we can guess for the next observation based on what we have seen in
the past.  We are not aware of any anomaly detection performance
bounds in non-stationary or adversarial settings prior to our work.

\subsection{Notation}

We will follow the following notational conventions. ``Basic'' sets
will be denoted by sans-serif uppercase letters, e.g.,~$\cU,\cX,\cZ$,
while classes of sets and functions will be denoted by script letters,
e.g.,~$\cC,\cF,\cG$. Given a set $\cX$, we will denote by $\cX^k$ the
$k$-fold Cartesian product of $\cX$ with itself and by $x^k$ a
representative $k$-tuple from $\cX^k$. The set of all (one-sided)
infinite sequences over $\cX$ will be denoted by $\cX^\infty$, and a
representative element will be written in boldface as $\bd{x} =
(x_1,x_2,\ldots)$. The interior of a set $\cU$ will be denoted by
$\Int \cU$. The standard Euclidean inner product between two vectors
$u,v \in \reals^m$ will be denoted by $\ave{u,v}$.

\section{Preliminaries}
\label{sec:prelims}

This section is devoted to setting up the basic terminology and machinery to be used throughout the paper. This includes background information on online convex programming (Section~\ref{ssec:OCP}) and on exponential families (Section~\ref{ssec:exp_families}).

\subsection{Online convex programming}
\label{ssec:OCP}

The philosophy advocated in the present paper is that the tasks of
sequential probability assignment and threshold selection can both be
viewed as a {\em game} between two opponents, the Forecaster and the
Environment. The Forecaster is continually predicting changes in a
dynamic Environment, where the effect of the Environment is
represented by an arbitrarily varying sequence of convex cost
functions over a given feasible set, and the goal of the Forecaster is
to pick the next feasible point in such a way as to keep the
cumulative cost as low as possible. This is broadly formulated as the
problem of {\em online convex programming}, or OCP
\cite{Zin03,BarHazRak08,ABRT08}. An OCP problem with horizon $T$ is specified by a
convex feasible set $\cU \subseteq \reals^d$ and a family of convex
functions $\cF = \{ f : \cU \to \reals\}$, and is described as
follows:
\begin{algorithm}[h]
\caption{An abstract Online Convex Programming problem}
\begin{algorithmic}

\STATE The Forecaster picks an arbitrary initial point $\wh{u}_1 \in \cU$

\FOR{$t=1,2,\ldots,T$}
\STATE The Environment picks a convex function $f_t \in \cF$
\STATE The Forecaster observes $f_t$ and incurs the cost $f_t(\wh{u}_{t})$
\STATE The Forecaster picks a new point $\wh{u}_{t+1} \in \cU$
\ENDFOR
\end{algorithmic}
\end{algorithm}

\sloppypar \noindent{The total cost incurred by the Forecaster after
  $T$ rounds is given by $\sum^T_{t=1} f_t(\wh{u}_{t})$ (here and in
  the sequel, hats denote quantities selected by the Forecaster on the
  basis of past observations). At each time $t$, the Forecaster's move
  $\wh{u}_t$ must satisfy a causality constraint in that it may depend
  only on his past moves $\wh{u}^{t-1}$ and on the past functions
  $f^{t-1}$ selected by the Environment. Thus, the behavior of the
  Forecaster may be described by a sequence of functions $$ \su_t :
  \cU^{t-1} \times \cF^{t-1} \to \cU, \qquad t = 1,2,\ldots,T $$ so
  that $\wh{u}_t = \su_t(\wh{u}^{t-1},f^{t-1})$. We will refer to any
  such sequence $\mu^T = \{\su_t : \cU^{t-1} \times \cF^{t-1} \to \cU
  \}^T_{t=1}$ as a $T$-step {\em strategy} of the Forecaster.
  Informally, the goal of the Forecaster is to do almost as well as if
  he could observe the cost functions $f_1,\ldots,f_T$ all at
  once. For instance, we might want to minimize the difference between
  the actual cost incurred after $T$ rounds of the game and the
  smallest cumulative cost that could be achieved {\em in hindsight}
  using a single feasible point. To that end, given a strategy $\mu^T$
  and a cost function tuple $f^T$, let us define the {\em regret}
  w.r.t.\ a time-varying tuple $u^T = (u_1,\ldots,u_T) \in
  \cU^T$ \begin{align*}
    R_T(\su^T; f^T, u^T) &\deq \sum^T_{t=1} f_t(\wh{u}_t) - \sum^T_{t=1}f_t(u_t) \\
    &= \sum^T_{t=1} f_t(\su_t(\wh{u}^{t-1},f^{t-1})) - \sum^T_{t=1}
    f_t(u_t).  \end{align*} Then the goal would be to select a
  suitable restricted subset $\cC_T \subset \cU^T$ and design $\mu^T$
  to ensure that the worst-case regret \begin{align*}
    &\sup_{f^T \in \cF^T} \sup_{u^T \in \cC_T} R_T(\su^T; f^T, u^T) \\
    & \qquad \equiv \sup_{f^T \in \cF^T} \left\{ \sum^T_{t=1}
      f_t(\wh{u}_t) - \inf_{u^T \in \cC_T} \sum^T_{t=1}
      f_t(u_t)\right\} \end{align*} is sublinear in $T$.  (When $u_t =
  u$ for all $t$ and for some $u \in \cU$, we will write the regret as
  $R_T(\mu^T;f^T,u)$, and the second $\sup$ in the worst-case regret
  would be over all $u \in
  \cU$.)  

Note that it is often convenient to think of a comparison tuple $u^T \in \cC_T$ as a strategy of the form $\mu_t : \cU^{t-1} \times \cF^{t-1} \to \{u_t\}$, i.e., $u_t$ does not depend on the previous points or cost functions, but only on the time index $t$. This allows us to speak of comparison classes of {\em strategies}; however, to avoid confusion, we will always use the notation $\su^T$ (possibly with subscripts) to distinguish the Forecaster's observation-driven strategy from a comparison strategy $u^T$, which may be time-varying but is always observation-independent. This is similar to the distinction made in control theory between {\em closed-loop} (or {\em feedback}) policies and {\em open-loop} policies \cite{KumVar86}: a closed-loop policy is a sequence of {\em functions} for selecting the next control signal based on the past control signals and past observations, while an open-loop policy is a sequence of control {\em signals} fixed in advance. From this point of view, the regret pertains to the difference in cumulative costs between a feedback policy $(\mu^T)$ and the best open-loop policy in some reference class $\cC$.

\begin{remark}[Hannan consistency] {\em More generally, we can consider unbounded-horizon strategies $\bd{\su} = \{\su_t : \cU^{t-1} \times \cF^{t-1} \to \cU \}$. Then, given a {\em comparison class} $\cC \subset \cU^\infty$ of open-loop strategies, the design goal is to ensure that \begin{equation}\label{eq:OCP_objective} R_T(\bd{\su}; \cC) \deq \sup_{\bd{f} \in \cF^\infty} \sup_{\bd{u} \in \cC} R_T(\su^T; f^T, u^T) = o(T).  \end{equation} Any strategy $\bd{\su}$ that achieves (\ref{eq:OCP_objective}) over a comparison class $\cC$ is said to be {\em Hannan-consistent} w.r.t.\ $\cF$ and $\cC$; see the text of Cesa-Bianchi and Lugosi \cite{CesLug06} for a thorough discussion. One important comparison class is composed of all {\em static} (or {\em constant}) sequences, i.e., all $\bd{u} \in \cU^\infty$ such that $u_1 = u_2 = \ldots$. This class, which we will denote by $\cC_{\text{stat}}$, is in one-to-one correspondence with the feasible set $\cU$, so $$ R_T(\bd{\su}; \cC_{\text{stat}}) = \sup_{\bd{f} \in \cF^\infty} \sup_{u \in \cU} R_T(\mu^T; f^T, u), $$ and we will also denote this worst-case regret by $R_T(\bd{\su}; \cU)$.}
\end{remark}

\subsection{The mirror descent procedure}
\label{sssec:mirror_descent}

A generic procedure for constructing OCP strategies is inspired by the
so-called {\em method of mirror descent}
\cite{NemYud83,BecTeb03,NJLS09}. In the context of OCP, the rough idea
behind mirror descent is as follows. At time $t$ the Forecaster
chooses the point \begin{equation}\label{eq:MDA} \wh{u}_{t+1} =
  \argmin_{u \in \cU} \Big[ \eta_t \ave{g_t(\wh{u}_t),u} +
  D(u,\wh{u}_t) \Big], \end{equation} where $g_t(\wh{u}_t)$ is an
arbitrary subgradient\footnote{A subgradient of a convex function $f :
  \cU \to \reals$ at a point $u \in \Int \cU$ is any vector $g \in
  \reals^d$, such that $$ f(v) \ge f(u) + \ave{ g, v - u } $$ holds
  for all $v \in \Int \cU$ \cite{HirLem01}.} of $f_t$ at $\wh{u}_t$,
$D(\cdot,\cdot) \ge 0$ is some measure of proximity between points in
$\cU$, and $\eta_t > 0$ is a (possibly time-dependent) regularization
parameter. The intuition behind (\ref{eq:MDA}) is to balance the
tendency to stay close to the previous point against the tendency to
move in the direction of the greatest local decrease of the cost. The
key feature of mirror descent methods is that they can be flexibly
adjusted to the geometry of the feasible set $\cU$ through judicious
choice of the proximity measure $D(\cdot,\cdot)$. In particular, when
$\cU$ is the canonical parameter space of an exponential family, a
good proximity measure is the Kullback--Leibler divergence. The
general measures of proximity used in mirror descent are given by the
so-called {\em Bregman divergences} \cite{Bre67,CenZen97}. Following
\cite{CesLug06}, we introduce them through the notion of a {\em
  Legendre function}:
\begin{definition}
  Let $\cU \subseteq \reals^d$ be a nonempty set with convex
  interior. A function $F : \cU \to \reals$ is called {\em Legendre}
  if it is: \begin{enumerate} \item Strictly convex and continuously
    differentiable throughout $\Int \cU$; \item {\em Steep} (or {\em
      essentially smooth}) --- that is, if $u_1,u_2,\ldots \in \Int
    \cU$ is a sequence of points converging to a point on the boundary
    of $\cU$, then $\| \nabla F(u_i) \| \to \infty$ as $i \to \infty$,
    where $\| \cdot \|$ denotes any norm.\footnote{Since all norms on
      finite-dimensional spaces are equivalent, it suffices to
      establish essential smoothness in a particular norm, say the
      usual $\ell_2$ norm.}  \end{enumerate} The {\em Bregman
    divergence} induced by $F$ is the nonnegative function $D_F : \cU
  \times \Int \cU \to \reals$, given by $$ D_F(u,v) \deq F(u) - F(v) -
  \ave{ \nabla F(v), u - v }, \forall u \in \cU, v \in \Int
  \cU.  $$\end{definition}

\noindent For example, if $\cU = \reals^d$, then $F(u) = (1/2) \| u
\|^2$, where $\| \cdot \|$ is the Euclidean norm, is Legendre, and
$D_F(u,v) = (1/2) \| u - v \|^2$. In general, for a fixed $v \in \Int
\cU$, $D_F(\cdot,v)$ gives the tail of the first-order Taylor
expansion of $F(\cdot)$ around $v$.

We now present the general mirror descent scheme for OCP, where we also allow the
possibility of restricting the feasible points to a closed, convex
subset $\cS$ of $\Int \cU$.

\begin{algorithm}[H] \caption{A Generic Mirror Descent
    Strategy for OCP} \label{alg:OCP} \begin{algorithmic} \REQUIRE A
    Legendre function $F : \cU \to \reals$; a decreasing sequence of
    strictly positive {\em step sizes} $\{\eta_t\}$
    \\
    \STATE The
    Forecaster picks an arbitrary initial point $\wh{u}_1 \in \cS$
    \FOR{$t=1,2,...$} \STATE Observe the cost function $f_t \in \cF$
    \STATE Compute a subgradient $g_t(\wh{u}_t)$ at $\wh{u}_t$
    \STATE Output
    $$
    \wh{u}_{t+1} = \argmin_{u \in \cS} \big[ \eta_t \ave{g_t(\wh{u}_t),u} + D_F(u,\wh{u}_t)\big]
    $$ \ENDFOR \end{algorithmic} \end{algorithm}

\noindent In the case when $\cU = \reals^d$ and $F(\cdot) = (1/2) \| \cdot
\|^2$, the above algorithm reduces to the standard projected
subgradient scheme \begin{align*}
  \td{u}_{t+1} &= \wh{u}_t - \eta_t g_t(\wh{u}_t) \\
  \wh{u}_{t+1} &= \argmin_{u \in \cS} \left\| u - \td{u}_{t+1}
  \right\|.  \end{align*} 
  
The name ``mirror descent" comes from the following equivalent form of Algorithm~\ref{alg:OCP}. Consider the {\em Legendre--Fenchel dual} of $F$
\cite{HirLem01,BoyVan04}:
$$
F^*(z) \deq \sup_{u \in \cU} \left\{
  \ave{u, z} - F(u) \right\}.
$$
Let $\cU^*$
denote the image of $\Int \cU$ under the gradient mapping $\nabla
F$: $\cU^* = \nabla F(\Int \cU)$. An important fact is that the gradient mappings $\nabla F$ and
$\nabla F^*$ are inverses of one another
\cite{BecTeb03,CesLug06,NJLS09}: \begin{align*}
  \left.  \begin{array}{l}
      \nabla F^* (\nabla F(u)) = u \\
      \nabla F ( \nabla F^* (w)) = w \end{array} \right\} \qquad
  \forall u \in \Int \cU, w \in \Int \cU^* \end{align*} 
  Following
\cite{CesLug06}, we may refer to the points in $\Int \cU$ as the {\em
  primal points} and to their images under $\nabla F$ as the {\em dual
  points}.
Then, for each $t$, the computation of $\wh{u}_{t+1}$ in Algorithm~\ref{alg:OCP} can be implemented as follows:
\begin{enumerate}
\item Compute $\xi_t = \nabla F (\wh{u}_t)$
\item Perform dual update $\td{\xi}_{t+1} = \xi_t - \eta_t g_t(\wh{u}_t)$
\item Compute $\td{u}_{t+1} = \nabla F^*(\td{\xi}_{t+1})$
\item Perform projected primal update:
$$
\wh{u}_{t+1} = \argmin_{u \in \cS} D_F(u,\td{u}_{t+1})
$$
\end{enumerate}
  The name ``mirror descent'' reflects the fact
that, at each iteration, the current point in the primal space is
mapped to its ``mirror image'' in the dual space; this is followed by
a step in the direction of the negative subgradient, and then the new
dual point is mapped back to the primal
space. In the context of mirror descent schemes, the Legendre
function $F$ is referred to as the {\em potential function}.
  
The
following lemma (see, e.g.,~Lemma~2.1 in \cite{NJLS09}) is a key
ingredient in bounding the regret of the mirror descent
strategy:

\begin{lemma}\label{lm:basic_regret_bound} Fix an arbitrary norm $\| \cdot \|$ on $\reals^d$ and suppose that, on the set $\cS$, the Legendre potential $F$ is {\em strongly convex} w.r.t.\ $\| \cdot \|$ with parameter $\alpha > 0$, i.e., for any $u,u' \in \Int \cS$,
\begin{align}\label{eq:SC_potential}
F(u') \ge F(u) + \ave{\nabla F(u),u'-u} + \frac{\alpha}{2} \| u - u' \|^2.
\end{align} 
Then for any $u \in \cS$ and any $t$, we have the bound
\begin{align}
& D_F(\wh{u}_{t+1},u) \le D_F(\wh{u}_t,u) \nonumber\\
& \qquad + \eta_t\ave{g_t(\wh{u}_t),u-\wh{u}_t} + \frac{\eta^2_t}{2\alpha} \| g_t(\wh{u}_t) \|^2_*,\label{eq:one_time_step}
\end{align}
where $\| v \|_* \deq \sup \{ \ave{u,v} : \| u \| \le 1 \}$ is the norm {\em dual} to $\| \cdot \|$.
\end{lemma}
\begin{remark}{\em The Euclidean norm $\| u \| = (u^2_1 + \ldots + u^2_d)^{1/2}$ is dual to itself, $\| \cdot \|_* = \| \cdot \|$.} 
\end{remark}
\noindent The possibility of attaining sublinear regret using mirror descent hinges on the availability of a suitable strongly convex Legendre potential. Typically, the choice of the potential function is influenced by the geometry of the underlying set $\cU$; the reader is invited to consult the papers by Beck and Teboulle \cite{BecTeb03} and by Nemirovski {\em et al.} \cite{NJLS09} for many examples.

The above Lemma~\ref{lm:basic_regret_bound} is central to our proofs of all the regret bounds presented in the sequel. Thus, even though the overall flavor of the proofs is similar to what can be found in the OCP literature \cite{Zin03,CesLug06,BarHazRak08,ABRT08}, we feel that the use of Lemma~\ref{lm:basic_regret_bound} (instead of the more usual arguments exploiting the primal-dual form of MD and the ``three-point formula" for Bregman divergences \cite[Lemma~11.1]{CesLug06}) leads to simpler and shorter arguments and allows us to seamlessly tie together many different settings (e.g., both static and time-varying comparison classes). The reader may also wish to consult \cite{RRY}, where Lemma~\ref{lm:basic_regret_bound} is used to analyze adaptive closed-loop control schemes based on mirror descent with data-driven selection of the step size.

\subsection{Background on exponential
  families} \label{ssec:exp_families} This section is devoted to a
brief summary of the basics of 
exponential families; Amari and Nagaoka \cite{AmaNag00} or Wainwright
and Jordan 
\cite{WaiJor08} give more details.  

We assume that the observation space $\cX$ is equipped with a
$\sigma$-algebra $\cB$ and a $\sigma$-finite measure $\nu$ on
$(\cX,\cB)$. Given a positive integer $d$, let $\phi : \cX \to
\reals^d$ be a measurable function, and let $\phi_k$,
$k=1,2,\ldots,d$, denote its components:
$$ \phi(x) = \big( \phi_1(x),\ldots,\phi_d(x) \big)^T.  $$ Let
$\Theta$ 
be the set of all $\theta \in \reals^d$ such that 
$$ \int_\cX \exp \big\{{\ave{\theta,\phi(x)}}\big\} d\nu(x) <
+\infty.  $$ 
We then have the following definition: \begin{definition} The set
$\cP(\phi)$ of probability distributions on $(\cX,\cB)$ parametrized
by $\theta \in \Theta$, such that the probability density function of
each $P_\theta \in \cP(\phi)$ w.r.t.\ the measure $\nu$ can be
expressed as $$ p_\theta(x) = \exp \big\{ \ave{\theta,\phi(x)} -
\Phi(\theta) \big\}, $$ where $$ \Phi(\theta) \deq \log \int_\cX \exp
\big\{\ave{\theta,\phi(x)} \big\} d\nu(x), $$ is called an {\em
exponential family} with {\em sufficient statistic} $\phi$. The
parameter $\theta \in \Theta$ is called the {\em natural parameter}
of $\cP(\phi)$, and the set $\Theta$ is called the {\em natural
parameter space.} The function $\Phi$ is called the {\em log
partition function}.\footnote{This usage comes from statistical
physics.}\end{definition} 

We will denote by $\expect_\theta[\cdot]$ the expectation w.r.t.\ $P_\theta$: \begin{align*}
  \expect_\theta [ g(X) ] &= \int_\cX g(x) dP_\theta(x) \\
  &= \int_\cX g(x) \exp \big\{\ave{\theta,\phi(x)} - \Phi(\theta)
  \big\} d\nu(x).  \end{align*} \begin{example} \label{ex:bernoulli} {\em The simplest
    example is the Bernoulli distribution. In this case,
    $\cX=\{0,1\}$, $\nu$ is the counting measure, $\phi(x) = x$, and
    $\Phi(\theta) = \log[1+\exp(\theta)]$. The natural parameter space
    $\Theta$ is the entire real line. Under this parametrization, we
    have $p_\theta(X=1)
    =e^\theta/(1+e^\theta)$.\hfill$\square$} \end{example} \begin{example}[The
  Ising model]\label{ex:ising}{\em Consider an undirected graph $G =
    (V,E)$ and associate with each vertex $\alpha \in V$ a binary
    random variable $X_\alpha \in \{-1,+1\}$. In this case, $\cX =
    \{-1,+1\}^V$, $\nu$ is the counting measure, and we have the
    following density for the random variable $X = (X_\alpha : \alpha
    \in V) \in \{-1,+1\}^V$: \begin{align} p_\theta(x) = \exp \left\{
        \sum_{\alpha \in V} \theta_\alpha x_\alpha +
        \sum_{(\alpha,\beta) \in E} \theta_{\alpha\beta} x_\alpha
        x_\beta - \Phi(\theta) \right\}, \label{eq:ising} \end{align}
    where $\theta$ is the collection of $d=|V|+|E|$ real parameters
    $(\theta_\alpha : \alpha \in V)$ and $(\theta_{\alpha\beta} :
    (\alpha,\beta) \in E)$. The log partition function is $$
    \Phi(\theta) = \log \sum_{x \in \cX} \exp \left\{ \sum_{\alpha \in
        V} \theta_\alpha x_\alpha + \sum_{(\alpha,\beta) \in E}
      \theta_{\alpha\beta}x_\alpha x_\beta \right\}.  $$ The
    sufficient statistic $\phi$ is given by the functions $\phi_\alpha
    : x \mapsto x_\alpha, \alpha \in V$ and $\phi_{\alpha\beta} : x
    \mapsto x_\alpha x_\beta, (\alpha,\beta) \in E$. Since
    $\Phi(\theta)$ is finite for any choice of $\theta$, we have
    $\Theta = \reals^d$, with the components of $\theta$ appropriately
    ordered.\hfill$\square$}
\end{example}
    
\begin{example}[Gaussian
  Markov random fields]\label{ex:gaussian_MRF} {\em Again, consider an
    undirected graph $G = (V,E)$.  For notational convenience, let us
    number the vertices as $V=\{1,\ldots,p\}$. A Gaussian Markov
    random field (MRF) on $G$ is a multivariate Gaussian random
    variable $X = (X_1,\ldots,X_p)^T \in \reals^p$, where the
    covariates $X_\alpha$ and $X_\beta$ are independent if
    $(\alpha,\beta) \not\in E$.  Then $\cX = \reals^p$ and $\nu$ is
    the Lebesgue measure. The distribution of $X$ can be written
    exactly as in
    \eqref{eq:ising}, 
the log partition function $\Phi(\theta)$ is
finite only if the $p \times p$ matrix $\Gamma \deq
[\theta_{\alpha\beta}]^p_{\alpha,\beta=1}$ is negative definite
($\Gamma \prec 0$), so that the parameter space is $\Theta = \{((\theta_1,\ldots,\theta_p)^T,\Gamma) \in \reals^p \times \reals^{p \times p} : \Gamma
\prec 0, \Gamma = \Gamma^T
\}$.\hfill$\square$} \end{example}

\subsection{General properties of
exponential families}

The motivation behind our use of exponential
families is twofold: (1) They form a sufficiently rich class of
parametric statistical models (which includes Markov random fields with
pairwise interactions) and can be used to describe
co-occurrence data, visual scene snapshots, biometric records, and
many other categorical and numerical data types. Moreover, they can be
used to approximate many nonparametric classes of probability
densities \cite{BarShe91}.  (2) The negative log-likelihood function
is {\em convex} in the natural parameter and {\em affine} in the
sufficient statistic. This structure permits the use of OCP.  We will
need the following facts about exponential families (proofs can be
found in the references listed at the beginning of
Section~\ref{ssec:exp_families}): \begin{enumerate} \item The log
  partition function $\Phi : \reals^d \to \reals \cup \{-\infty,+\infty\}$ is lower semicontinuous on
  $\reals^d$ and infinitely differentiable on $\Theta$.  \item The
  derivatives of $\Phi$ at $\theta$ are the cumulants of the random
  vector $\phi(X) = (\phi_1(X),\ldots,\phi_d(X))$ when $X \sim
  p_\theta$. 
In particular, \begin{align*}
    \nabla \Phi(\theta) &= \big( \expect_\theta \phi_1(X), \ldots, \expect_\theta \phi_d(X) \big)^T\\
    \nabla^2 \Phi(\theta)& = \left[ \text{Cov}_\theta( \phi_i(X),
      \phi_j (X)) \right]^d_{i,j=1}.  \end{align*} Thus, the Hessian
  $\nabla^2 \Phi(\theta)$, being the covariance matrix of the vector
  $\phi(X)$, is positive semidefinite, which implies that
  $\Phi(\theta)$ is a convex function of $\theta$. In particular,
  $\Theta$, which, by definition, is the essential domain of $\Phi$,
  is convex.  \item $\Phi(\theta)$ is {\em steep} (or {\em essentially
    smooth}): if $\{\theta_n\} \subset \Theta$ is a sequence
  converging to some point $\theta$ on the boundary of $\Theta$, then
  $\| \nabla \Phi(\theta_n) \| \to + \infty$ as $n \to \infty$.  
    \item The
  relative entropy (Kullback--Leibler divergence) between
  $p_{\theta_1}$ and $p_{\theta_2}$ in $\cP(\phi)$, defined as
  $D(p_{\theta_1} \| p_{\theta_2}) = \int_\cX p_{\theta_1} \log
  (p_{\theta_1}/p_{\theta_2}) d\nu$, can be written
  as \begin{equation}\label{eq:KL_divergence} D(p_{\theta_1} \|
    p_{\theta_2}) = \Phi(\theta_2) - \Phi(\theta_1) - \ave{\nabla
      \Phi(\theta_1), \theta_2 - \theta_1} \end{equation} From now on,
  we will use the shorthand $D(\theta_1 \|
  \theta_2)$.  \end{enumerate} From these properties, it follows that
$\Phi : \Theta \to \reals$ is a Legendre function, and that the
mapping $D_\Phi : \Theta \times \Int \Theta \to \reals$, defined by
$D_\Phi(\theta_1,\theta_2) = D(\theta_2 \| \theta_1)$, is a Bregman
divergence.

\section{Filtering: sequential
  probability assignment in the presence of
  noise} \label{sec:prob_assign_with_noise}
  
  The first ingredient of
\FHTAGN\ is a strategy for assigning a likelihood (or belief)
$p_t(\cdot|z^{t-1})$ to the clean symbol $x_t$ based on the past noisy
observations $z^{t-1}$. Alternatively, we can think of the following
problem: if $x_t$ is the actual clean symbol that has been generated
at time $t$, then our likelihood $p_t \equiv p_t(x_t|z^{t-1})$, though
well-defined, is not accessible for observation. Thus, we would like
to {\em estimate} it via some estimator $\wh{p}_t$, which will depend
on the actual observed noisy symbol $z_t$, as well as on the
previously obtained estimates $\wh{p}^{t-1} =
(\wh{p}_1,\ldots,\wh{p}_{t-1})$. In the field of signal processing,
problems of this kind go under the general heading of {\em filtering};
this term refers to any situation in which it is desired, at each time
$t$, to obtain an estimate of some clean unobservable quantity {\em
  causally} based on noisy past observations.

\subsection{Sequential probability assignment: general formulation}

\subsubsection{Noiseless observations}

Let us first consider the noiseless case, i.e., $z_t = x_t$ for all
$t$.  Elements of an arbitrary sequence $\bd{x} = (x_1,x_2,\ldots) \in
\cX^\infty$ are revealed to us one at a time, and we make no
assumptions on the law that generates $\bd{x}$. At each time $t =
1,2,\ldots$, before $x_t$ is revealed, we have to assign a probability
density $\wh{p}_t$ (w.r.t.\ a fixed dominating measure $\nu$) to the
possible values of $x_t$. When $x_t$ is revealed, we incur the {\em
  logarithmic loss} $-\log \wh{p}_t(x_t)$ (the choice of this loss
function is standard and is motivated by information-theoretic
considerations; cf.\ the survey paper by Merhav and Feder
\cite{MerFed98} for more details). Let $\cD$ denote the set of all
valid probability densities w.r.t.\ $\nu$. Then the sequential
probability assignment can be represented by a sequence $\bd{\sp}$ of
mappings $\sp_t : \cX^{t-1} \to \cD$, so that
$$
\wh{p}_t = \sp_t(x^{t-1}), \qquad \text{or } \wh{p}_t(x_t) = [\sp_t(x^{t-1})](x_t).
$$
We refer to any such sequence of probability assignments $\bd{\sp}$ as
a {\em prediction strategy}. Since the probability assignment
$\wh{p}_t$ is a function of the past observations $x^{t-1}$, we may
also view it as a conditional probability density
$\wh{p}_t(\cdot|x^{t-1})$. In the absence of specific probabilistic
assumptions on the generation of $\bd{x}$, it is appropriate to view
$$
\wh{P}_t(A|x^{t-1}) \deq \int_A \wh{p}_t(x|x^{t-1}) d\nu(x)
$$
as our {\em belief}, based on the past observations $x^{t-1}$, that
the next observation $x_t$ will lie in a measurable set $A \subseteq
\cX$. Another way to think about $\bd{\sp}$ is as a sequence of joint
densities
$$
\wh{p}^T(x^T) = \prod^T_{t=1} \wh{p}_t(x_t|x^{t-1}), \qquad T = 1,2,\ldots.
$$

In an individual-sequence setting, the performance of a given
prediction strategy is compared to the best performance achievable on
$\bd{x}$ by any strategy in some specified comparison class $\cC$
\cite{MerFed98,CesLug06}. Any such comparison strategy is also
specified by a sequence of conditional densities $p_t(x_t|x^{t-1})$ of
$x_t$ given $x^{t-1}$. Suppose first that the horizon $T$ is fixed in
advance. Given a prediction strategy $\bd{\sp} =
\{\sp_t\}^\infty_{t=1}$, we can define the {\em regret} w.r.t.\
$\bd{p} = \{p_t\} \in \cC$ after $T$ time steps as
\begin{align}
& R_T(\sp^T; x^T,p^T) \nonumber\\
&\quad\deq \sum^T_{t=1} \log \frac{1}{[\sp_t(x^{t-1})](x_t)} - \sum^T_{t=1} \log \frac{1}{p_t(x_t|x^{t-1})} \nonumber \\
&\quad=  \sum^T_{t=1} \log \frac{1}{\wh{p}_t(x_t|x^{t-1})} - \sum^T_{t=1} \log \frac{1}{p_t(x_t|x^{t-1})}.
  \label{eq:regret}
\end{align}
As before, the distinction between $\sp_t$ and $\wh{p}_t$ is that the
former is a mapping of $\cX^{t-1}$ into the space of probability
densities $\cD$, while the latter is the image of $x^{t-1}$ under
$\sp_t$.

\subsubsection{Noisy observations}

We are interested here in a more difficult problem, namely sequential
probability assignment {\em in the presence of noise}. That is, instead of
observing the ``clean'' symbols $x_t \in \cX$, we receive ``noisy''
symbols $z_t \in \cZ$ (where $\cZ$ is some other observation
space). We assume that the noise is stochastic, memoryless and
stationary. In other words, at each time $t$, the noisy observation
$z_t$ is given by $z_t = N(x_t,r_t)$, where $\{r_t\}$ is an i.i.d.\
random sequence and $N(\cdot,\cdot)$ is a fixed deterministic
function. There are two key differences between this and the noiseless
setting described earlier, namely:
\begin{enumerate}
\item The prediction strategy now consists of mappings $\wh{\sp}_t : \cZ^{t-1} \to \cD$, where, at each time $t$, $\wh{p}_t(\cdot|z^{t-1}) = \wh{\sp}_t(z^{t-1})$ is the conditional probability density we assign to the {\em clean} observation $x_t$ at time $t$ given the past {\em noisy} observations $z^{t-1}$.
\item We cannot compute the true incurred log loss $-\log \wh{p}_t(x_t|z^{t-1})$.
\end{enumerate}
We are interested in sequential prediction, via the beliefs $\wh{\sp}_t(z^{t-1}) = \wh{p}_t(\cdot|z^{t-1})$, of the next clean symbol $x_t$ given the past noisy observations $z^{t-1}$. We assume, as before, that the clean sequence $\bd{x}$ is an unknown individual sequence over $\cX$. Moreover, under our noise model the noisy observations $\{z_t\}$ are conditionally independent of one another given $\bd{x}$.

\subsection{Probability assignment in an exponential family via OCP-based filtering}
\label{ssec:OCP_exp_families}

We will now show that if the comparison class $\cC$ consists of
product distributions lying in an {\em exponential family} with
natural parameter $\theta \in \reals^d$, then we can use OCP to design
a scheme for sequential probability assignment from noisy data. The
use of OCP is made possible by the fact that, in an exponential
family, the (negative) log likelihood is a {\em convex} function of
the natural parameter and an {\em affine} function of the sufficient
statistic.

Recall from Section~\ref{ssec:exp_families} that a $d$-dimensional exponential family consists of probability densities of the form $p_\theta(x) =  e^{\ave{ \theta, \phi(x) } - \Phi(\theta)}$, where the parameter
$\theta$ lies in a convex subset of $\reals^d$. We will consider prediction strategies of the
form
\begin{align}\label{eq:exp_fam_strategy}
\wh{\sp}_t (z^{t-1}) = p_{\wh{\theta}_t}(\cdot), \qquad t=1,2,\ldots
\end{align}
where $\wh{\theta}_t$ is a function of the past noisy observations $z^{t-1}$. The log loss function in this particular case takes the form
$$
-\log \wh{p}_t(x_t) = -\ave{ \wh{\theta}_t,\phi(x_t) } + \Phi(\wh{\theta}_t) , \qquad x \in \cX.
$$
Thus, the regret relative to any comparison strategy $\bd{p}_{\bd{\theta}}$ induced by a parameter sequence $\bd{\theta} = \{\theta_t\} \in \Theta^\infty$ via $p_t(\cdot|x^{t-1}) = p_{\theta_t}(\cdot)$ can be written as
\begin{align*}
R_T(\wh{\sp}^T; x^T, p^T_{\bd{\theta}}) 
&= \sum^T_{t=1} \log \frac{1}{p_{\wh{\theta}_t}(x_t)} - \sum^T_{t=1} \log \frac{1}{p_{\theta_t}(x_t)} \\
&= \sum^T_{t=1} \left[ \ell(\wh{\theta}_t,x_t) - \ell(\theta_t,x_t) \right],
\end{align*}
where we have defined the function
$$
\ell(\theta,x) \deq -\ave{ \theta, \phi(x) } + \Phi(\theta).
$$
Because the log partition function $\Phi$ is convex, the function $\theta \mapsto \ell(\theta,x)$ is convex for every fixed $x \in \cX$. Therefore, if the observations were noiseless, i.e.,\ $z_t = x_t$ for all $t$, then mirror descent could have been used to design an appropriate strategy of the form (\ref{eq:exp_fam_strategy}). As we will show next, this approach also works in the noisy case, provided an unbiased estimator of $\phi(x)$ based on the noisy observation $z$ is available. In fact, our results in the noisy setting contain the noiseless setting as a special case.

Let us fix an exponential family $\cP(\phi)$. We will consider the comparison class consisting of product
distributions, where each marginal belongs to a certain subset
of $\cP(\phi)$. Specifically, let $\Lambda$ be a closed, convex subset of $\Theta$.  We take $\cC$ to consist of prediction
strategies $\bd{p}_{\bd{\theta}}$, where $\bd{\theta} =
(\theta_1,\theta_2,\ldots) \in \Lambda^\infty$ ranges over all infinite sequences over
$\Lambda$, and each $\bd{p}_{\bd{\theta}}$ is of the product form $p_{\theta_1} \otimes p_{\theta_2} \otimes \ldots$, i.e.,
\begin{equation}
p_{t,\bd{\theta}}(\cdot|x^{t-1}) = p_{\theta_{t}}(\cdot), \qquad x^{t-1} \in \cX^{t-1}, t = 1,2,\ldots.
\label{eq:product_strategies}
\end{equation}
In other words, each prediction strategy in $\cC$ corresponds
to a time-varying product density whose marginals belong to $\{ p_\theta :
\theta \in \Lambda\}$. From now on, we will use the term ``strategy'' to refer to an infinite sequence $\bd{\theta} \in \Lambda^\infty$; the corresponding object $\bd{p}_{\bd{\theta}}$ will be implied.

Consider the noisy observation model $z_t = N(x_t,r_t)$, where
$\{r_t\}$ is the i.i.d.\ noise process and $N(\cdot,\cdot)$ is a known
deterministic function.  We make the following assumption:

\begin{assumption} There exists a function $h : \cZ \to \reals^d$, such that $\expect [h(z)|x] = \phi(x)$, where the expectation is taken w.r.t.\ the noise input $r$. In other words, $h(z)$ is an unbiased estimator of the sufficient statistic $\phi(x)$.
\end{assumption}
\noindent Here, the conditional notation $\expect [h(z)|x]$ is
shorthand for the fact that the expectation is taken w.r.t.\ the
common distribution $Q$ of $r_1,r_2,\ldots$, while keeping the clean
input $x$ fixed:
\begin{align}\label{eq:xcond}
\expect[h(z)|x] = \int h(N(x,r)) dQ(r).
\end{align}

\begin{example} {\em In Example~\ref{ex:bernoulli}, $x \in \{0,1\}^d$
    and $\phi(x) = x$, $r \in \{0,1\}^d$, where each component $r(i)$
    is a $\text{Bernoulli}(p)$ random variable independent of
    everything else $(p < 1/2)$, and
\begin{equation}
z(i) = x(i) \oplus r(i), \qquad i = 1,\ldots,d.
\label{eq:binary_noise}
\end{equation}
In other words, every component of $z$ is independently related to the corresponding component of $x$ via a {\em binary symmetric channel} (BSC) with crossover probability $p$ \cite{CovTho06}. Then an unbiased estimator for $\phi(x) \equiv x$ is given by $h(z) = (h_1(z),\ldots,h_d(z))^T$,
where
$$
h_i(z) = \frac{z(i)-p}{1-2p}, \qquad i = 1,\ldots,d.
$$}
\end{example}

\begin{example} {\em Consider the Ising model from Example~\ref{ex:ising} and suppose that each $x_\alpha$, $\alpha \in V$, is independently corrupted by a BSC with crossover probability $p$. Then an unbiased estimator for $\phi(x)$ is given by
\begin{align*}
h_\alpha(z) &= \frac{z_\alpha - p}{1-2p}, \qquad \alpha \in V \\
h_{\alpha\beta}(z) &= \frac{z_\alpha - p}{1-2p} \cdot \frac{z_\beta - p}{1-2p}, \qquad (\alpha,\beta) \in E
\end{align*}
so that we have $\expect[h_\alpha(z)|x] = \phi_\alpha(x) = x_\alpha$ and $\expect[h_{\alpha\beta}(z)|x] = \phi_{\alpha\beta}(x) = x_\alpha x_\beta$.\hfill $\square$}
 \end{example}

 \begin{example}{\em Consider the Gaussian MRF from Example~\ref{ex:gaussian_MRF} and suppose that each $x_\alpha$, $\alpha \in V$, is independently corrupted by an additive white Gaussian noise (AWGN) channel with noise variance $\sigma^2$ \cite{CovTho06}. In other words, $z = (z_\alpha : \alpha \in V)$, where, for each $\alpha \in V$, $z_\alpha = x_\alpha + r_\alpha$ with $r_\alpha \sim \text{Normal}(0,\sigma^2)$ independent of all other $r_\beta, \beta \neq \alpha$. Then an unbiased estimator for $\phi(x)$ is given by
 \begin{align*}
h_\alpha(z) = z_\alpha, h_{\alpha\beta}(z) &= \begin{cases}
z^2_\alpha - \sigma^2, & \alpha = \beta \\
z_\alpha z_\beta, & \alpha \neq \beta
\end{cases}, \qquad \alpha,\beta \in V
 \end{align*}
 so that $\expect[h_\alpha(z)|x] = x_\alpha$ and
 $\expect[h_{\alpha\beta}(z)|x] = x_\alpha x_\beta$.\hfill$\square$}
\end{example}

\noindent By virtue of our Assumption~1, the {\em filtering loss}
$$
\lf(\theta,z_t) \deq  -  \ave{ \theta,h(z_t)
} + \Phi(\theta)
$$
is an unbiased estimator of the true log loss $\ell(\theta,x_t)$ for any $\theta \in \Theta$:
\begin{align*}
\expect\Big[\lf(\theta,z_t)\Big|x_t\Big] &= -\ave{\theta,\expect[h(z_t)|x_t]} + \Phi(\theta) \\
&= - \ave{\theta,\phi(x_t)} + \Phi(\theta) = \ell(\theta,x_t).
\end{align*}
This leads to the following prediction strategy:

\begin{algorithm}[H]
\caption{Sequential Probability Assignment via Noisy Mirror Descent}
\label{alg:OCP_with_noise}
\begin{algorithmic}
\REQUIRE A closed, convex set $\Lambda \subset \Theta$; a decreasing sequence of strictly positive step sizes $\{\eta_t\}$

\STATE Initialize with $\wh{\theta}_1 \in \Lambda$

\FOR{$t=1,2,...$}
    \STATE Acquire new noisy observation $z_t$

   \STATE Compute the filtering loss $\lf_t(\wh{\theta}_t) = -\ave{\wh{\theta}_t, h(z_t)} + \Phi(\wh{\theta}_t)$

    \STATE Output
    $$
    \wh{\theta}_{t+1} = \argmin_{\theta \in \Lambda} \left[ \eta_t \ave{\nabla \lf_t(\wh{\theta}_t),\theta} + D(\wh{\theta}_t \| \theta) \right]
    $$

\ENDFOR
\end{algorithmic}
\end{algorithm}
\noindent This induces the following sequential probability assignment strategy:
\begin{subequations}\label{eq:obj2}
\begin{align}
&\sp_t = p_{\wh{\theta}_t}(\cdot) \\
&\wh{\theta}_{t+1} = \argmin_{\theta \in \Lambda} \left[\Ave{ \theta, \nabla \Phi(\wh{\theta}_t) - h(z_t) } +
\frac{1}{\eta_t}D(\wh{\theta}_t \| \theta)\right].
\end{align}
\end{subequations}
For the reader's convenience, Table~\ref{tbl:corr_OCP} shows the
correspondence between the objects used in
Algorithm~\ref{alg:OCP_with_noise} and the generic mirror descent
strategy, i.e.,~Algorithm~\ref{alg:OCP}. 

\begin{table*}[ht]
\begin{center}
{\small
\begin{tabular}{|l||c|c|}
\hline
& Generic MD & Algorithm~\ref{alg:OCP_with_noise} \\
\hline
\hline
Convex set & $\cU$ & $\Theta$ \\
\hline
Cost functions & $f_t$ & $\wh{\ell}(\cdot,z_t) \equiv - \ave{\cdot,h(z_t)} + \Phi(\cdot)$ \\
\hline
Project onto & $\cS$ & $\Lambda$ \\
\hline
Legendre potential & $F$ & $\Phi$ \\
\hline
Bregman divergence & $D_F(\cdot,\cdot)$ & $D_\Phi(\cdot,\cdot) \equiv D(\cdot \| \cdot)$ \\
\hline
\end{tabular}
}
\end{center}
\caption{\label{tbl:corr_OCP}Correspondence between the generic mirror
  descent (MD) and Algorithm~\ref{alg:OCP_with_noise}.} 
\end{table*}

This approach has the following features:
\begin{enumerate}
\item The geometry of exponential families leads to a natural choice
  of the Legendre potential and the corresponding Bregman divergence
  to be used in the mirror-descent updates, namely the log partition
  function $\Phi$ and the Kullback--Leibler divergence $D(\cdot \|
  \cdot)$. 
\item The optimization at each time can be computed using only the
  current noisy observation $z_t$ and the probability density $\wh{p}_t$ estimated at the
  previous time; it is not necessary to keep all observations in
  memory to ensure strong performance.
\end{enumerate}
Azoury and Warmuth \cite{AzoWar01} proposed and analyzed an algorithm
similar to (\ref{eq:obj2}) in the setting of online density
estimation over an exponential family. However, they did not consider
noisy observations and only proved regret bounds for a couple of
specific exponential families. One of the contributions of the present
paper is to demonstrate that minimax (logarithmic) regret bounds
against static strategies can be obtained for a {\em general}
exponential family, subject to mild restrictions on the feasible set
$\Lambda \subseteq \Theta$. This provides an answer to the question
posed by Azoury and Warmuth about whether it is possible to attain
logarithmic regret for a general exponential family.

\subsection{Regret bounds for OCP-based filter}
\label{ssec:regret_bounds}

We will now establish the following bounds on the expected regret of Algorithm~\ref{alg:OCP_with_noise}:
\begin{enumerate}
\item If the comparison class $\cC$ consists of {\em static} strategies $\theta_1 = \theta_2 = \ldots$ over $\Lambda$, then, under certain regularity conditions on $\Lambda$ and with properly chosen step sizes $\{\eta_t\}$,  the expected regret of the strategy in (\ref{eq:obj2}) will be $O(\log T)$.
\item Given a strategy $\bd{\theta}$ and a time horizon $T$, define the {\em variation} of $\bd{\theta}$ from $t=1$ to $t=T$ as
\begin{equation}
V_T(\bd{\theta}) \deq \sum^T_{t=1} \| \theta_t - \theta_{t+1} \|,
\label{eq:variation}
\end{equation}
where $\| \cdot \|$ is taken to be the $\ell_2$ norm for concreteness. If the comparison class $\cC$ consists of all time-varying strategies $\bd{\theta} = \{\theta_t\}$ over $\Lambda$, then, under certain regularity conditions on $\Lambda$ and with properly chosen step sizes $\{\eta_t\}$, the expected regret of the algorithm in  (\ref{eq:obj2}) will be $ O\big( (V_T(\bd{\theta})+1) \sqrt{T} \big)$.
 \end{enumerate}
The expectation in both cases is taken w.r.t.\ the noise process $\{r_t\}$. Moreover, in the absence of noise (i.e., $z_t = x_t$ for all $t$), the above regret bounds will hold for all observation sequences $\bd{x}$.

We will bound the regret of Algorithm~\ref{alg:OCP_with_noise} in two
steps. In the first step, we will obtain bounds on the regret computed
using the filtering losses $\lf(\cdot,\cdot)$ that hold for {\em any}
realization of the noisy sequence $\bd{z} = \{z_t\}$. In the second
step, we will use a martingale argument along the lines of Weissman
and Merhav \cite{WeissmanMerhav_PredictionBinaryITIT01} to show that
the expected ``true'' regret is bounded by the expected filtering
regret.

\subsubsection{Regret bounds for the filtering loss}
\label{sssec:filtering_loss_regret}

We will consider time-varying strategies of the form (\ref{eq:product_strategies}), where the set $\Lambda$ is restricted in the following way. Given a positive constant $H > 0$, define the set
$$
\Theta_H \deq \Big\{ \theta \in \Theta : \nabla^2 \Phi(\theta) \succeq 2H I_{d \times d} \Big\},
$$
where $I_{d \times d}$ denotes the $d\times d$ identity matrix, and the matrix inequality $A \succeq B$ denotes the fact that $A-B$ is positive semidefinite. Note that the Hessian $\nabla^2 \Phi(\theta)$ is equal to
$$
J(\theta)
\deq -\expect_\theta[\nabla^2_\theta \log p_\theta(X)],
$$
which is the Fisher information matrix at $\theta$
\cite{ClaBar90,AmaNag00,WaiJor08}. In other words, $\Theta_H$ consists
of all parameter vectors $\theta \in \Theta$, for which the
eigenvalues of the Fisher information matrix over $\Lambda$ are
bounded from below by $2H$.  Yet another motivation for our definition
of $\Theta_H$ comes from the fact that $\nabla^2 \Phi(\theta)$ is the
covariance matrix of the random vector $\phi(X) =
(\phi_1(X),\ldots,\phi_d(X))^T$ when $X \sim P_\theta$. Thus, a strictly positive
uniform lower bound on the eigenvalues of this covariance matrix
implies that any coordinate of $\phi(X)$ is sufficiently
``informative" about (or correlates well with) the remaining
coordinates.  We will assume in the sequel that $\Lambda$ is any
closed, convex subset of $\Theta_H$.
 
For any strategy $\btheta = \{\theta_t\}$, define the cumulative true and estimated losses
\begin{align*}
L_{\btheta,T}(x^T) &\deq \sum_{t=1}^T \ell(\theta_t,x_t),\\
\Lf_{\btheta,T}(z^T) &\deq \sum_{t=1}^T \lf(\theta_t,z_t)
\end{align*}
and the difference
\begin{align*}
\Df_{\btheta,T}(x^T,z^T) &\deq L_{\btheta,T}(x^T) -
\Lf_{\btheta,T}(z^T) \\
&= \sum_{t=1}^T \ave{ \theta_t, h(z_t)-\phi(x_t) }.
\end{align*}
When $\bd{\theta}$ is a static strategy corresponding to $\theta \in \Lambda$, we will write $L_{\theta,T}(x^T)$, $\Lf_{\theta,T}(z^T)$, and $\Df_{\theta,T}(x^T,z^T)$. We first establish a logarithmic regret bound against static strategies in $\Lambda$. The theorem below improves on our earlier result from \cite{raginsky_OCP_isit09}:

\begin{theorem}[Logarithmic regret against static
  strategies]\label{thm:log_regret} Let $\Lambda$ be any closed,
  convex subset of $\Theta_H$, and let $\wh{\bd{\theta}} =
  \{\wh{\theta}_t \}$ be the sequence of parameters in $\Lambda$
  computed from the noisy sequence $\bd{z} = \{z_t\}$ using the OCP
  procedure shown in Algorithm~\ref{alg:OCP_with_noise} with step
  sizes $\eta_t = 1/t$. Then, for any $\theta \in \Lambda$, we have
\begin{align}
\Lf_{\bhtheta,T}(z^T) &\leq
\Lf_{\theta,T}(z^T)  + \frac{(K(z^T)+M)^2}{ H}(\log T + 1),
\label{eq:const_regret}
\end{align}
where
\begin{align*}
K(z^T) \deq \frac{1}{2} \max_{1 \le t \le T}  \| h(z_t) \| \text{ and } M \deq \frac{1}{2} \max_{\theta \in \Lambda} \| \nabla \Phi(\theta) \|.
\end{align*}
\end{theorem}

\begin{IEEEproof} Appendix~\ref{app:proof_log_regret}.
\end{IEEEproof}

With larger step sizes $\eta_t = 1/\sqrt{t}$, it is possible to compete against {\em time-varying} strategies $\bd{\theta} = \{\theta_t\}$, provided the variation is sufficiently slow:

\begin{theorem}[Regret against time-varying
  strategies]\label{thm:root_regret} Again, let $\Lambda$ be any
  closed, convex subset of $\Theta_H$. Let $\wh{\bd{\theta}}$ be the
  sequence of parameters in $\Lambda$ computed from the noisy sequence
  $\bd{z} = \{z_t\}$ using the OCP procedure shown in
  Algorithm~\ref{alg:OCP_with_noise} with step sizes $\eta_t =
  1/\sqrt{t}$. Then, for any sequence $\bd{\theta} = \{\theta_t\}$
  over $\Lambda$, we have
\begin{align*}
  \Lf_{\wh{\bd{\theta}},T}(z^T) &\le \Lf_{\bd{\theta},T}(z^T) + 4M\sqrt{T}V_T(\bd{\theta}) \\
  & \qquad \qquad + \frac{(K(z^T)+M)^2}{H}(2 \sqrt{T}-1),
\end{align*}
where $K(z^T)$ and $M$ are defined as in Theorem~\ref{thm:log_regret}, and $V_T(\bd{\theta})$ is defined in (\ref{eq:variation}).
\end{theorem}
\begin{IEEEproof} Appendix~\ref{app:proof_root_regret}.\end{IEEEproof}

\begin{remark} {\em Regret bounds against a dynamically changing
    reference strategy have been derived in a variety of contexts,
    including prediction with expert advice with finitely many
    time-varying experts \cite{HerWar98}, sequential linear prediction
    \cite{HerWar01}, general online convex programming \cite{Zin03},
    and sequential universal lossless source coding (which is
    equivalent to universal prediction with log-loss)
    \cite{Wil96,ShaMer99}. It is useful to compare the result of
    Theorem~\ref{thm:root_regret} with some of these bounds. The
    results of Herbster and Warmuth \cite{HerWar01} and Zinkevich
    \cite{Zin03} assume fixed and known horizon $T$. Furthermore,
    Herbster and Warmuth \cite{HerWar01} assume that the loss function
    is of {\em subquadratic type} (see, e.g.,~Section~11.4 in
    \cite{CesLug06}) and use a different scale-dependent notion of
    regret and a carefully adjusted time-independent step size. On the
    other hand, Zinkevich \cite{Zin03} uses a constant step size
    $\eta$ and obtains a regret bound of the form $O( V_T/\eta +
    T\eta)$, where the constants implicit in the $O(\cdot)$ notation
    depend on the diameter of the feasible set and on the maximum norm
    of the (sub)gradient of the loss function. It is not hard to
    modify the proof of our Theorem~\ref{thm:root_regret} to obtain a
    similar regret bound in our case, including the constants. The
    results of Shamir and Merhav \cite{ShaMer99} (which extend the
    work of Willems \cite{Wil96}) deal with universal prediction of
    piecewise-constant memoryless sources under log-loss. They propose
    two types of algorithms --- those with linearly growing per-round
    computational complexity, and those with fixed per-round
    complexity. For the first type, and with $V_T = O(1)$, they obtain
    $O(\log T)$ regret (which is optimal \cite{Mer93}); for the second
    type, again with $V_T = O(1)$, they develop two schemes, one of
    which attains $O(\log T)$ regret for certain sources with ``large"
    jumps and $O(T \log \log T / \log T)$ in general, while the other
    always achieves $O(\sqrt{T \log T})$ regret. Since our algorithms
    have fixed per-letter complexity, it is clear that we cannot
    achieve the optimal $O(\log T)$ regret; however, our $O(\sqrt{T})$
    regret in the $V_T = O(1)$ case compares favorably against the
    second fixed-complexity algorithm of \cite{ShaMer99}. Of course,
    the reason why our algorithms have fixed per-letter complexity
    comes from the special structure of the log-loss function for an
    exponential family, which maps the problem into an instance of OCP
    on the underlying parameter space.  }
\end{remark}

\subsubsection{Bounds on the expected true regret}
\label{ssec:expected_regret}

We now proceed to establish regret bounds on
$L_{\btheta,T}(x^T)$. The bounds of Theorems~\ref{thm:log_regret} and \ref{thm:root_regret} reflected how close our cumulative loss might be to that of a competing strategy on noisy data. We now show that our proposed strategy ensures that the {\em expected} cumulative loss on the {\em unobserved} clean data is close to that of competing strategies. First, we need the following lemma, which is
similar to Lemma 1 in
\cite{WeissmanMerhav_PredictionBinaryITIT01}:

\begin{lemma}\label{lem:martingale} Let $\bd{r} = \{r_t\}$ be the
  i.i.d.\ observation noise process. For each $t$, let $\cR_t$ denote
  the $\sigma$-algebra generated by $r_1,\ldots,r_t$. Let $\btheta =
  \{\theta_t\}$ be a sequence of probability assignments, such that
  each $\theta_t = \theta_t(z^{t-1})$. Then, for any individual
  sequence $\bd{x} = \{x_t\}$,
$\left\{\Df_{\btheta,t}(x^t,z^t), \cR_t\right\}$ is a martingale, and
so  $\expect \Lf_{\btheta,T}(z^T) = \expect L_{\btheta,T}(x^T)$ for each $T$. The expectation is conditional on the underlying clean sequence $\bd{x}$, cf.\ \eqref{eq:xcond}.
\end{lemma}
\begin{IEEEproof} Appendix~\ref{app:proof_martingale}.\end{IEEEproof}

This leads to regret bounds on the proposed OCP-based filter:

\begin{theorem}\label{thm:filter} Consider the setting of
  Theorem~\ref{thm:log_regret}. Then we have
\begin{align}
  & \expect [ L_{\wh{\bd{\theta}},T}(x^T)] \leq \inf_{\theta \in \Lambda} L_{\theta,T}(x^T) \nonumber\\
  & \qquad \qquad + \frac{\expect [ (K(z^T)+M)^2]}{H} (\log T + 1).
\label{eq:exp_log_regret}
\end{align}
Likewise, in the setting of Theorem~\ref{thm:root_regret}, we have
\begin{align}
  & \expect [ L_{\wh{\bd{\theta}},T}(x^T)  ] \nonumber \\
  & \qquad \leq \inf_{\btheta} \left[L_{\btheta,T}(x^T)  +  4 M\sqrt{T}V_T(\btheta)\right] \nonumber \\
  &\qquad \qquad + \frac{\expect[(K(z^T)+M)^2]}{H}(2\sqrt{T}-1),
\label{eq:exp_root_regret}
\end{align}
where the infimum is over all strategies $\bd{\theta}$ over $\Lambda$,
and the expectation is conditional on the underlying clean sequence
$\bd{x}$.
\end{theorem}
\proof We will only prove (\ref{eq:exp_root_regret}); the proof of
(\ref{eq:exp_log_regret}) is similar. Proceeding analogously to the
proof of Theorem 4 in \cite{WeissmanMerhav_PredictionBinaryITIT01}, we
have
\begin{align*}
  &\expect L_{\bhtheta,T}(x^T) = \expect \Lf_{\bhtheta,T}(z^T)\\
  &\leq \expect \Big\{
  \inf_{\bd{\theta}} \left[\Lf_{\btheta,T}(z^T) + 4 M\sqrt{T}V_T(\bd{\theta}) \right] \\
  & \qquad \qquad  + \frac{\expect [K(z^T)+M)^2]}{H} (2\sqrt{T}-1)\Big\}\\
  &\leq  \inf_{\btheta} \left[ \expect \Lf_{\btheta,T}(z^T) +  4 M\sqrt{T} V_T(\btheta) \right] \\
  & \qquad \qquad  + \frac{\expect[(K(z^T)+M)^2]}{H}(2\sqrt{T}-1)  \\
  &= \inf_{\btheta} \left[ L_{\btheta,T}(x^T)  + 4 M\sqrt{T} V_T(\btheta) \right] \\
  & \qquad \qquad + \frac{\expect[(K(z^T)+M)^2]}{H}(2\sqrt{T}-1) ,
\end{align*}
where the first step follows from Lemma~\ref{lem:martingale}, the
second  from Theorem~\ref{thm:root_regret}, the third  from the fact that $\expect \inf [\cdot] \le \inf \expect [\cdot]$, and the last from Lemma~\ref{lem:martingale} and the fact that the
expectation is taken with respect to the distribution of $z_t|x_t$.
\endproof

\begin{remark} {\em In the usual regret notation, the bounds of Theorem~\ref{thm:filter} can be written as follows:
$$
 \expect R_T(\sp^T; x^T, \theta) \le \frac{\expect[(K(z^T) + M)^2]}{H}(\log T +1 )
 $$
and
\begin{align*}
&\expect R_T(\sp^T; x^T, \theta^T) \le 4 M \sqrt{T} V_T(\bd{\theta}) \\
& \qquad \qquad \qquad \qquad + \frac{\expect[(K(z^T) + M)^2]}{H}(2\sqrt{T}-1).
\end{align*}}
\end{remark}

\subsubsection{Minimax optimality and Hannan consistency}

Finally, we make a few comments regarding minimax optimality and Hannan consistency of the strategies described in this section.

Recall the OCP game described in Section~\ref{ssec:OCP}. During each round $t = 1,2,\ldots,T$, the Forecaster plays a point $\wh{u}_t \in \cU$, the Environment responds with a convex function $f_t \in \cF$, and the Forecaster incurs the cost $f_t(\wh{u}_t)$. The Forecaster's goal is to keep the cumulative cost $\sum^T_{t=1} f_t(\wh{u}_t)$ as low as possible. Let us suppose, moreover, that the Environment is antagonistic in that it tries to choose the functions $f_t$, so that the current cumulative cost $\sum^t_{s=1} f_s(\wh{u}_s)$ is as high as possible, given the past moves of the Forecaster, $\wh{u}^t$, and the past cost functions $f^{t-1}$. To allow the Environment more freedom, we assume that the cost function at time $t$ is selected from a set $\cF_t$ which may depend on the current move $\wh{u}_t$.  With this in mind, let us define, following Abernethy et al.~\cite{ABRT08}, the {\em minimax value} of the game as
\begin{align}\label{eq:minimax_value}
& R^*_T(\cU,\cF^T) = \inf_{u_1 \in \cU} \sup_{f_1 \in \cF_1} \ldots \inf_{u_T \in \cU} \sup_{f_T \in \cF_T} \nonumber\\
& \qquad \qquad \qquad \left\{ \sum^T_{t=1} f_t(u_t) - \inf_{u \in \cU} \sum^T_{t=1} f_t(u) \right\}.
\end{align}
In words, $R^*_T(\cU,\cF^T)$ is the worst-case regret of an {\em optimal} strategy for the Forecaster. The order of the infima and the suprema in (\ref{eq:minimax_value}) reflects the order of the moves and the causality restrictions. Thus, the Forecaster's move at time $t$ may depend only on his moves and on the cost functions revealed at times $s=1,\ldots,t-1$; the Environment's cost function at time $t$ may depend only on the Forecaster's moves at times $s=1,\ldots,t$ and on the cost functions at times $s=1,\ldots,t-1$. Then we have the following bounds:

\begin{theorem}[\cite{ABRT08}]\label{thm:ABRT_minimax} Suppose that $\cU \subset \reals^d$ is compact and convex, and there exist some constants $G,\sigma > 0$ such that  at each time $t$ the functions in $\cF_t$ satisfy the following conditions:
$$
\cF_t = \left\{ f : \| \nabla f(\wh{u}_t) \| \le G, \nabla^2 f \succeq \sigma I_{d \times d} \right\}
$$
Then
$$
\frac{G^2}{2\sigma} \log (T+1) \le R^*_T(\cU,\cF^T) \le \frac{G^2}{2\sigma} (\log T + 1).
$$
\end{theorem}

We can particularize this result to our case. Consider the setting of Theorem~\ref{thm:log_regret} in the noiseless regime: $x_t = z_t, \forall t$. Let us fix a constant $K > 0$ and let $\cU = \Lambda$ and
\begin{align*}
\cF_1 &= \ldots = \cF_T = \Big\{ f_x = \ell(\cdot,x) \\
&\qquad \qquad = - \ave{\cdot,\phi(x)} + \Phi(\cdot) : x \in \cX, \| \phi(x) \| \le 2K \Big\}.
\end{align*}
Then, by hypothesis, each $f_x \in \cF_t$ satisfies
$$
\| \nabla f_x(\theta) \| \le \| \phi(x) \| + \| \nabla \Phi(\theta) \| \le 2(K+M).
$$
Moreover, each $f_x \in \cF_t$ satisfies $\nabla^2 f_x(\theta) = \nabla^2 \Phi(\theta) \succeq 2H I_{d\times d}$ at every $\theta \in \Lambda$. Thus, applying Theorem~\ref{thm:ABRT_minimax} with $G = 2(K+M)$ and $\sigma = 2H$, we get
\begin{align*}
\frac{(K+M)^2}{H}\log (T+1) &\le R^*_T(\Lambda,\cF^T) \\
&\le \frac{(K+M)^2}{H} (\log T + 1).
\end{align*}
On the other hand, with these assumptions we have $K(z^T) = K(x^T) = K$, and so our regret bound from Theorem~\ref{thm:log_regret} is of the form
$$
R_T \le \frac{(K+M)^2}{H} (\log T + 1),
$$
and thus we attain the minimax value $R^*_T(\Lambda,\cF^T)$.

We can also establish a weak form of Hannan consistency for the OCP filters of Theorems~\ref{thm:log_regret} and \ref{thm:root_regret}.  Let $\cG$ denote the set of all sequences $\bd{x} \in \cX^\infty$, such that
\begin{align}\label{eq:good_sequence}
\expect[(K(z^T) + M)^2] = o(T/\log T).
\end{align}
For example, if $h$ and $\phi$ have uniformly bounded norms, then $\cG \equiv \cX^\infty$. Then  the filter of Theorem~\ref{thm:log_regret} satisfies
$$
\limsup_{T \to \infty} \frac{1}{T} \sup_{\theta \in \Lambda} \expect [R_T(\sp^T; x^T, \theta) ] = 0
$$
for any $\bd{x} \in \cG$. As for the filter of Theorem~\ref{thm:root_regret}, consider any set $\cC$ of all time-varying strategies $\bd{\theta} \in \Lambda^\infty$, such that
$$
\sup_{\bd{\theta} \in \cC} V_T(\bd{\theta}) = o(\sqrt{T}).
$$
Then
\begin{align}\label{eq:hannan_constant}
\limsup_{T \to \infty} \frac{1}{T} \sup_{\bd{\theta} \in \cC} \expect [ R_T(\sp^T; x^T, \theta^T) ] = 0
\end{align}
for any $\bd{x} \in \cG$. For example, consider the set $\cC$ of all piecewise constant strategies $\bd{\theta} \in \Lambda^\infty$, such that $k_T$, the maximum number of switches in any $\theta^{T+1}$, satisfies $k_T = o(\sqrt{T})$. Assume also that $\Lambda$ is compact. Then for any $\bd{\theta} \in \cC$ we have
\begin{align}\label{eq:hannan_dynamic}
\sup_{\bd{\theta} \in \cC} V_T(\bd{\theta}) &= \sup_{\bd{\theta} \in \cC} \sum^T_{t=1} \| \theta_{t+1} - \theta_t \|\nonumber\\
& \le \diam (\Lambda) \cdot k_T \nonumber\\
&= o(\sqrt{T}),
\end{align}
so we have Hannan consistency.

Stronger bounds that hold {\em uniformly} over the choice of the observation sequence $\bd{x}$ are also possible, provided the convergence in (\ref{eq:good_sequence}) is uniform; in other words, if we have a set $\bar{\cG} \subseteq \cX^\infty$, such that
$$
\limsup_{T \to \infty} \sup_{\bd{x} \in \bar{\cG}} \frac{\expect[(K(z^T) + M)^2] \log T}{T} = 0,
$$
then the convergence in (\ref{eq:hannan_constant}) and (\ref{eq:hannan_dynamic}) is uniform in $\bd{x} \in \bar{\cG}$.

\section{Hedging: sequential threshold selection for anomaly detection}
\label{sec:anomaly_detection}

In the preceding section we have shown how to perform {\em filtering}, i.e., how to assign a belief $\wh{p}_t = \wh{p}_t(z^t)$ to the clean symbol $x_t$, such that
$$
\sum^T_{t=1} \expect\Bigg[\log \frac{1}{\wh{p}_t(z^{t-1})}\Bigg|x^t\Bigg] \approx  \sum^T_{t=1} \log \frac{1}{p_t(x_t|x^{t-1})},
$$
 where $p_t(\cdot|\cdot)$, $t=1,\ldots,T$, is the optimal sequence of conditional probability assignments, under a (possibly time-varying) exponential family model, for the entire clean observation sequence $x^T$.

The second ingredient of \FHTAGN\ is {\em hedging}, i.e., sequential
adjustment of the threshold $\tau_t$, such that whenever $\zeta_t \deq
\zeta(\wh{p}_t) < \tau_t$, where $\zeta: \reals^+ \to \reals$ is a user-specified
monotonically increasing function, we flag $z_t$ as anomalous. 

\begin{remark} {\em Note
that this formulation is equivalent to sequentially setting a
threshold $\td{\tau}_t$ such that, whenever $\wh{p}_t < \td{\tau}_t$, we
flag $z_t$ as anomalous. Using the monotone transformation $\zeta$ allows
us to sidestep challenging numerical issues when $\wh{p}_t$ is very
small. We will elaborate on this point later.}
\end{remark}

In order to choose an
appropriate $\tau_t$, we rely on feedback from an
end user. Specifically, let the end user set the label $y_t$ as $1$ if
$z_t$ is anomalous and $-1$ if $z_t$ is not anomalous. However, since
it is often desirable to minimize human intervention and analysis of
each observation, we seek to limit the amount of feedback received. To
this end, two possible scenarios could be considered:
\begin{itemize}
\item At each time
$t$, the Forecaster randomly decides whether to request a label from
the end user. A label is requested with probability that may depend on
$f_t$ and $\tau_t$.
\item At each time $t$, the end-user
arbitrarily chooses whether to provide a label to the Forecaster; the
Forecaster has no control over whether or not it receives a label.
\end{itemize}
As
we will see, the advantage of the first approach is that it allows us
to bound the average performance over all possible choices of times at
which labels are received, resulting in stronger bounds. The advantage
of the second approach is that is may be more practical or convenient
in many settings. For instance, if an anomaly is by chance noticed by
the end user or if an event flagged by the Forecaster as anomalous is,
upon further investigation, determined to be non-anomalous, this
information is readily available and can easily be provided to the
Forecaster. In the sequel, we will develop performance bounds for both
of these regimes.

In both settings, we will be interested in the number of mistakes made by the Forecaster over $T$ time steps. At each time step $t$, let $\wh{y}_t$ denote the binary label output by the Forecaster, $
\wh{y}_t = \sgn(\tau_t - \zeta_t)$, where we define $\sgn(a) = - 1$ if $a \le 0$ and $+1$ if $a > 0$. The number of mistakes over $T$ time steps is given by
\begin{equation}
\sum^T_{t=1} 1_{\{ \wh{y}_t \neq y_t\}} \equiv \sum^T_{t=1} 1_{\{ \sgn(\tau_t - \zeta_t) \neq y_t \}}.
\label{eq:num_mistakes}
\end{equation}
For simplicity, we assume here that the time horizon $T$ is known in
advance. We would like to obtain regret bounds relative to any fixed
threshold $\tau \in [\tau_{\min},\tau_{\max}]$ that could be chosen in hindsight after having observed the entire
sequence of ($\zeta$-transformed) probability assignments $\{\zeta_t\}^T_{t=1}$ and feedback
$\{y_t\}^T_{t=1}$ (note that some $y_t$'s may be ``empty,'' reflecting
the lack of availability of feedback at the corresponding
times). Here, $\tau_{\min}$
and $\tau_{\max}$ are some user-defined minimum and maximum threshold levels. Ideally, we would like to bound
\begin{equation}
\sum^T_{t=1} 1_{\{ \sgn(\tau_t - \zeta_t) \neq y_t\}} - \inf_{\tau\in[\tau_{\min},\tau_{\max}]}\sum^T_{t=1} 1_{\{\sgn(\tau - \zeta_t) \neq y_t\}}.
\label{eq:ideal_regret}
\end{equation}
However, analyzing this expression is difficult owing to the fact that
the  function $\tau \mapsto  1_{\{ \sgn(\tau  - \zeta)  \neq y\}}$ is not
convex  in $\tau$.  To  deal with  this  difficulty, we  will use  the
standard technique of replacing the comparator loss with a convex {\em
  surrogate function}. A frequently used surrogate is the {\em hinge loss}
$$
\ell(s,y) \deq
(1-sy)_+,
$$
where $(\alpha)_+ = \max\{0,\alpha\}$. Indeed, for any $\zeta$,
$\tau$ and $y$ we have
$$
1_{\{ \sgn(\tau - \zeta) \neq y\}} \le 1_{\{
  (\tau - \zeta)y < 0 \}} \le \big(1 - (\tau-\zeta)y\big)_+.
  $$
  Thus, instead of
(\ref{eq:ideal_regret}), we will bound the ``regret''
\begin{equation}
R_T(\tau) \deq \sum^T_{t=1} 1_{\{ \wh{y}_t \neq y_t \}} - \sum^T_{t=1} \ell_t(\tau),
\label{eq:surr_regret}
\end{equation}
where $\ell_t(\tau)$ is shorthand for $\ell(\tau - \zeta_t, y_t)$. In the following, we show that it is possible to obtain $O(\sqrt{T})$ surrogate regret using a modified mirror descent (more precisely, projected subgradient descent) strategy. The modifications are necessary to incorporate feedback into the updates.

We point out that the algorithms underlying the hedging step may be used in conjunction with any other method for assigning beliefs to incoming observations; however, together with our OCP-based filtering, they result in a low-complexity anomaly detection system with provable performance guarantees.

\subsection{Anomaly detection with full feedback}
\label{ssec:full_feedback}

In order to obtain bounds on the surrogate regret (\ref{eq:surr_regret}), we first analyze the ideal situation in which the Forecaster always receives feedback. Let $\Pi(\cdot)$ denote the projection onto the interval $[\tau_{\min},\tau_{\max}]$:
$$
\Pi(\alpha) \deq \argmin_{\tau \in [\tau_{\min},\tau_{\max}]} (\tau - \alpha)^2.
$$
In this setting, the following simple algorithm, which is essentially the  perceptron algorithm (see, e.g.,~Chapter 12 of \cite{CesLug06}) with projections,  does the job:

\begin{algorithm}[H]
\caption{Anomaly detection with full feedback}
\label{alg:exact_labels}
\begin{algorithmic}
\STATE {\bf Parameters:} real
numbers $\eta > 0$, $\tau_{\min}<\tau_{\max}$
\STATE {\bf Initialize:} $\tau_1 = \tau_{\min}$
\FOR{$t=1,2,\ldots,T$}
\STATE Receive the estimated likelihood $\wh{p}_t$, set $\zeta_t = \zeta(\wh{p}_t)$
\STATE \textbf{if} {$\zeta_t < \tau_t$} \textbf{then} flag $z_t$ as an
anomaly: $\wh{y}_t = 1$ \textbf{else} let $\wh{y}_t = -1$ 
 \STATE Obtain feedback $y_t$
\STATE Let $\tau_{t+1} = \Pi\big(\tau_t + \eta y_t 1_{\{ \wh{y}_t \neq y_t\}}\big)$
\ENDFOR
\end{algorithmic}
\end{algorithm}

Intuitively, the idea is this: if the Forecaster correctly assigns the label $\wh{y}_t$ to $z_t$, then the threshold stays the same; if the Forecaster incorrectly labels a nominal observation ($y_t = -1$) as anomalous ($\wh{y}_t = 1$), then the threshold is lowered: $\tau_{t+1} \approx \tau_t-\eta$; if the Forecaster incorrectly labels an anomalous observation ($y_t = 1$) as nominal ($\wh{y}_t = -1$), then the threshold is raised: $\tau_{t+1} \approx \tau_t + \eta$. We also observe that the above algorithm is of a mirror descent type with the Legendre potential $F(u) = u^2/2$, with one crucial difference: the current threshold $\tau_t$ is updated only when the Forecaster makes a mistake. We obtain the following regret bound:

\begin{theorem}\label{thm:exact_regret} Fix a time horizon $T$ and consider the Forecaster acting according to Algorithm~\ref{alg:exact_labels} with parameter $\eta = (\tau_{\max} - \tau_{\min})/\sqrt{T}$. Then, for any $\tau \in [\tau_{\min},\tau_{\max}]$, we have
\begin{equation}
R_T(\tau) = \sum^T_{t=1} 1_{\{ \wh{y}_t \neq y_t\}} - \sum^T_{t=1} \ell_t(\tau) \le (\tau_{\max} - \tau_{\min})\sqrt{T}.
\label{eq:exact_regret}
\end{equation}
\end{theorem}

\begin{IEEEproof} Appendix~\ref{app:proof_exact_hedge}. \end{IEEEproof}

\subsection{Random, Forecaster-driven feedback times}
\label{ssec:random_feedback_times}

We can now address the problem of online anomaly detection when the
Forecaster has an option to query the end-user for feedback when the
Forecaster is not sufficiently confident about its own decision
\cite{CesLugSto05}. Consider the following {\em label-efficient}
Forecaster for anomaly detection using sequential probability
assignments: 
\begin{algorithm}[H]
\caption{Label-efficient anomaly detection}
\label{alg:label_eff}
\begin{algorithmic}
\STATE {\bf Parameters:} real
numbers $\eta > 0$, $\tau_{\min}<\tau_{\max}$
\STATE {\bf Initialize:} $\tau_1 = \tau_{\min}$
\FOR{$t=1,2,\ldots$}
\STATE Receive the estimated likelihood $\wh{p}_t$, set $\zeta_t = \zeta(\wh{p}_t)$
\STATE \textbf{if} {$\zeta_t < \tau_t$} \textbf{then} flag $z_t$ as an
anomaly: $\wh{y}_t = 1$ \textbf{else} let $\wh{y}_t = -1$ 
\STATE Draw a Bernoulli random variable $U_t$ such that
$\Pr[U_t = 1|U^{t-1}] = 1/(1+|\zeta_t - \tau_t|)$
\STATE \textbf{if} $U_t = 1$ \textbf{then} request feedback $y_t$ and
let $\tau_{t+1} = \Pi\left(\tau_t + \eta y_t 1_{\{\wh{y}_t \neq y_t
    \}}\right)$
\textbf{else} let $\tau_{t+1} = \tau_t$
\ENDFOR
\end{algorithmic}
\end{algorithm}

\noindent This algorithm is, essentially, the label-efficient perceptron (see Section 12.4 of \cite{CesLug06}) with projections, and it attains the following regret:
\begin{theorem}\label{thm:label_eff}
Fix a time horizon $T$ and consider the label efficient Forecaster run
with parameter $\eta = 1/\sqrt{T}$. Then
$$
\expect \left[\sum^T_{t=1} 1_{\{\wh{y}_t \neq y_t\}} \right] \le \sum^T_{t=1} \ell_t(\tau) + (\tau_{\max} - \tau_{\min}) \sqrt{T}.
$$
where the expectation is taken with respect to $\{U_t\}$.
\end{theorem}
\begin{remark}[Computational issues involving very small
  numbers]\label{rem:comp_issues}{\em In some applications, the beliefs $\wh{p}_t$ may be very small. (For instance,
    in the Enron example presented in the experimental results,
    $\wh{p}_t = O(e^{-100})$.) In such a case, we will
    have $\Pr[U_t=1|U^{t-1}] \approx 1$, and our anomaly detection
    engine will request feedback almost all the time. To avoid this
    situation, the monotone transformation $\zeta$ is applied to
    $\wh{p}_t$ before thresholding. For instance, one might consider $\zeta(s) = Cs$ or $\zeta(s)
    = C\log s$ for an appropriately chosen positive number $C$. Note
    that the choice of $\zeta$ changes the form of the surrogate loss
    function. Thus care must be taken when choosing $\zeta$ to ensure (a)
    it approximates the original comparator loss as accurately as
    possible, (b) a reasonable number of feedback requests are made,
    and (c) numerical underflow issues are circumvented.}
\end{remark}
\begin{IEEEproof} Appendix~\ref{app:proof_label_eff}. \end{IEEEproof}

\subsection{Arbitrary feedback times}
\label{ssec:arbitrary_feedback}

When labels cannot be requested by the Forecaster, but are instead
provided arbitrarily by the environment or end user, we use the
following algorithm to choose the threshold $\tau$ at each time $t$:
\begin{algorithm}[H]
\caption{Anomaly detection with arbitrarily spaced feedback}
\label{alg:anomaly_arbitrary}
\begin{algorithmic}
\STATE {\bf Parameters:} real
number $\eta > 0$, $\tau_{\min}<\tau_{\max}$
\STATE {\bf Initialize:} $\tau_1 = \tau_{\min}$
\FOR{$t=1,2,\ldots,T$}
\STATE Receive the estimated likelihood $\wh{p}_t$, set $\zeta_t = \zeta(\wh{p}_t)$
\STATE \textbf{if} {$\zeta_t < \tau_t$} \textbf{then} flag $z_t$ as an
anomaly: $\wh{y}_t = 1$ \textbf{else} let $\wh{y}_t = -1$ 
\STATE \textbf{if} feedback $y_t$ is provided \textbf{then} 
let $\tau_{t+1} = \Pi\big(\tau_t + \eta y_t 1_{\{\wh{y}_t \neq
  y_t\}}\big)$
 \textbf{else} let $\tau_{t+1} = \tau_t$
\ENDFOR
\end{algorithmic}
\end{algorithm}
Under arbitrary feedback, it is  meaningful to compare the performance of the Forecaster against a comparator $\tau$ only at those times when the feedback is provided. We then have the
following performance bound:
\begin{theorem}\label{thm:anomaly_arbitrary}
Fix a time horizon $T$ and consider the anomaly detection with
arbitrarily spaced feedback Forecaster run
with parameter $\eta = 1/\sqrt{T}$. Let $t_1,\ldots,t_m$ denote the time steps at which the Forecaster receives feedback, and let $\epsilon \deq m/T$. Then, for any $\tau \in [\tau_{\min},\tau_{\max}]$, we have
$$
\sum^m_{i=1} 1_{\{\wh{y}_{t_i} \neq y_{t_i}\}} \le \sum^m_{i=1} \ell_{t_i}(\tau) + \frac{(1+\epsilon)(\tau_{\max}-\tau_{\min})}{2}\sqrt{T}.
$$
\end{theorem}

\begin{IEEEproof} If we consider only the times $\{t_1,\ldots,t_m\}$, then we are exactly in the setting of Theorem~\ref{thm:exact_regret}. This observation leads to the bound
$$
\sum^m_{i=1} 1_{\{\wh{y}_{t_i} \neq y_{t_i}\}} \le \sum^m_{i=1} \ell_{t_i}(\tau) + \frac{1}{2\eta} \left[ (\tau_{\max} - \tau_{\min})^2 + \epsilon T \eta^2 \right].
$$
With the choice $\eta = (\tau_{\max} - \tau_{\min})/\sqrt{T}$, we get the bound in the theorem.\end{IEEEproof}

\subsection{Arbitrary horizon}

So far, we have considered the case when the horizon $T$ is known in
advance. However, it is not hard to modify the proofs of the results
of this section to accommodate the case when the horizon is not set
beforehand. The only change that is required is to replace the
learning rate $\eta = O(1/\sqrt{T})$ with a time-varying sequence
$\{\eta_t\}$, where $\eta_t = O(1/\sqrt{t})$. Then the regret bounds
remain the same, namely, $O(\sqrt{T})$, possibly with different
constants.

\section{Experimental results}

In this section, we demonstrate how our proposed anomaly detection
approach \mbox{\FHTAGN} performs on simulated and real data.  We
consider four numerical experiments. Experiments 1, 2a and 2b use simulated data. {Experiment 1} tests the
{filtering} component of \FHTAGN. {Experiments
  2a} and {2b} focus on the {hedging} component of \FHTAGN.
{Experiment 3} applies \FHTAGN \ to the Enron email database \cite{Enron:04}.

\noindent{\bf Experiment 1.}  
For this experiment, we generate simulated data by drawing from a temporally-evolving Bernoulli product density. In particular,  we first draw i.i.d.\ samples according to $x_t \sim \prod_{i=1}^{500}
{\rm Bernoulli}(\beta^*_{i,t})$, where $\beta^*_{i,t} \in [0,1]$ and $1 \le t \le 1000$. Our observations $z_t \in \{0,1\}^{500}$ are the noisy versions of $x_t$, where
each bit $x_t(j)$ is passed through a BSC with crossover probability $0.1$. The goal is to causally estimate $\{\beta^*_{i,t}\}$ from $\{z_t\}$. Let $\mu^*_t \deq (\beta^*_{1,t},\ldots,\beta^*_{500,t})$. We choose $\mu^*_t$ to be piecewise constant in time, with changes at 
$t=100$, $500$ and
$700$. In this setting, knowing $\mu^*_t$ allows us to compute
empirical regret with respect to the known data generation parameters,
and to compare it against the theoretical regret bound in
Theorem~\ref{thm:root_regret}. 

We apply Algorithm \ref{alg:OCP_with_noise} to the above data with the
learning rate set to $\eta_t = 1/\sqrt{t}$.  Figure \ref{fig:exp1_mu}(a)
illustrates the ground truth $\mu_t^*$ vs.\ the estimated parameter
$\wh{\mu}_t$. Figures \ref{fig:exp1_mu}(c)--(d) show that the log-loss exhibits
pronounced spikes at the jump times, and then subsides as the
Forecaster adapts to the new parameters. Note that the variance of the
log-loss is larger for the noisy case. Finally, Figure
\ref{fig:exp1_mu}(e) shows that the empirical per-round regret is well
below the theoretical bound in Theorem \ref{thm:root_regret}, with the
regret for the noisy case again slightly larger.

\begin{figure*}[t]
\centering
\begin{tabular}{cc}  
\footnotesize{True $\mu_t^*$} & \footnotesize{Estimated $\wh{\mu}_t$}\\
\includegraphics[height=1.4in]{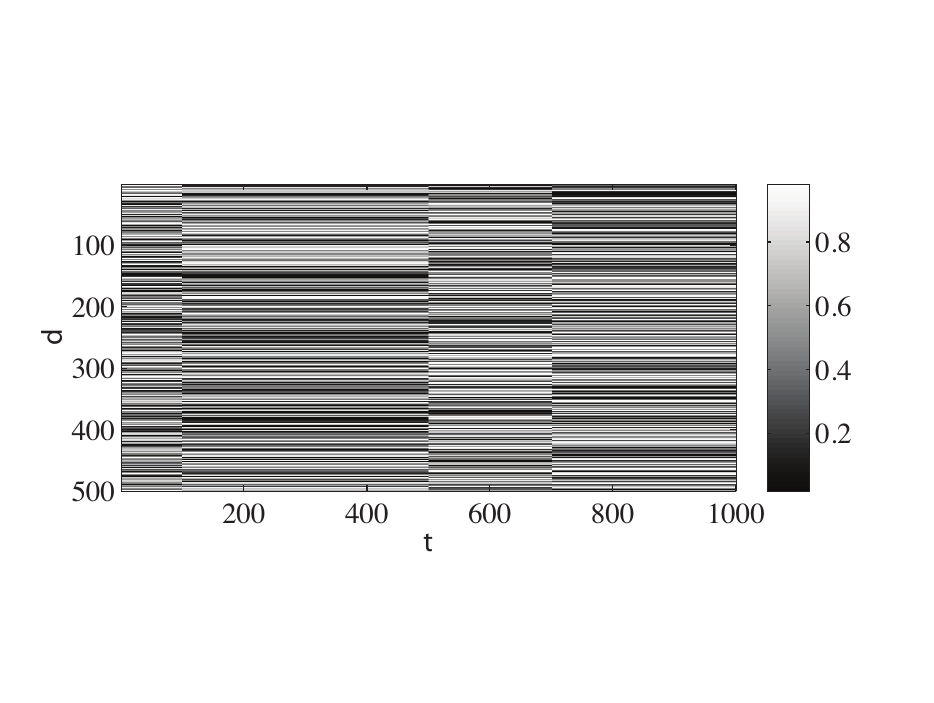}
&
\includegraphics[height=1.4in]{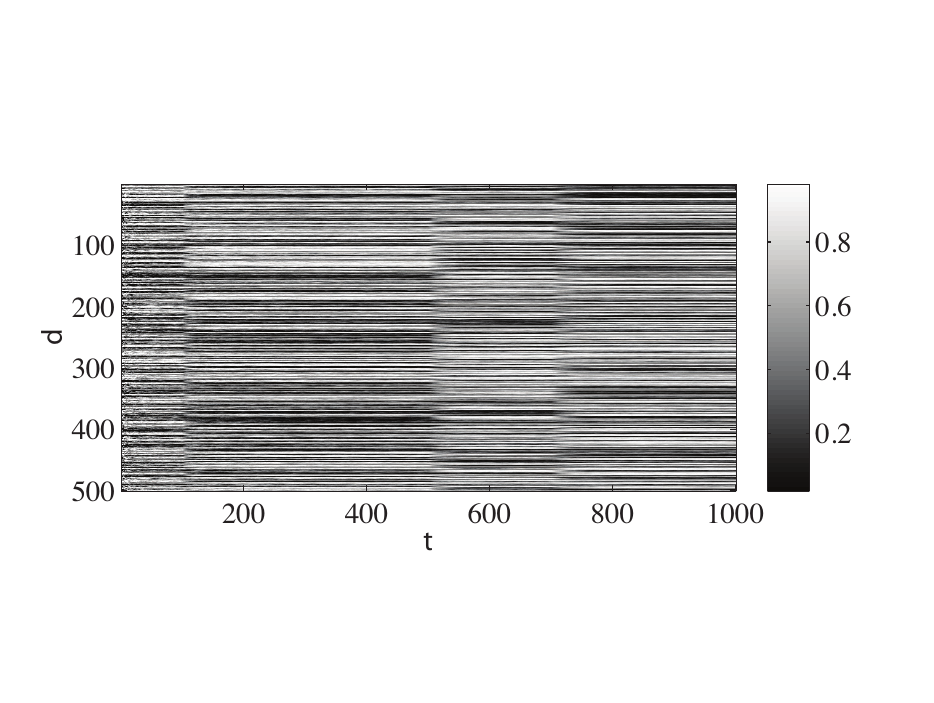}
\\
(a) & (b)
\end{tabular}
\begin{tabular}{ccc}
	\includegraphics[height=1.55in]{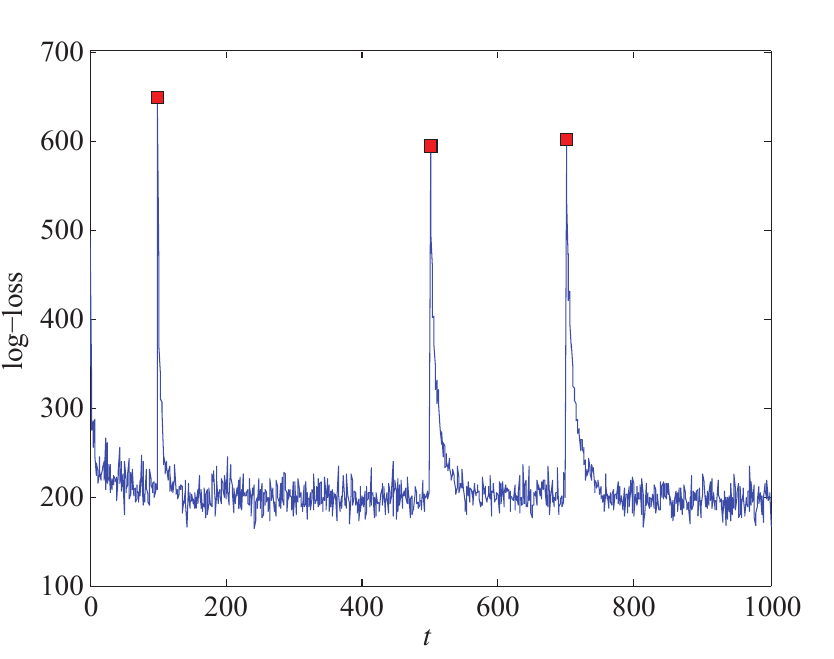} &
	\includegraphics[height=1.55in]{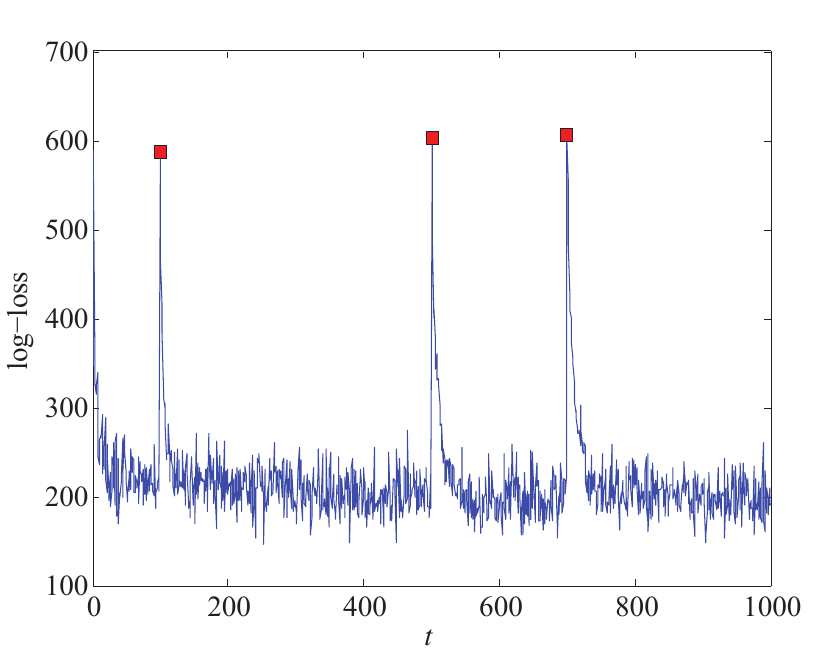} &
	\includegraphics[height=1.55in]{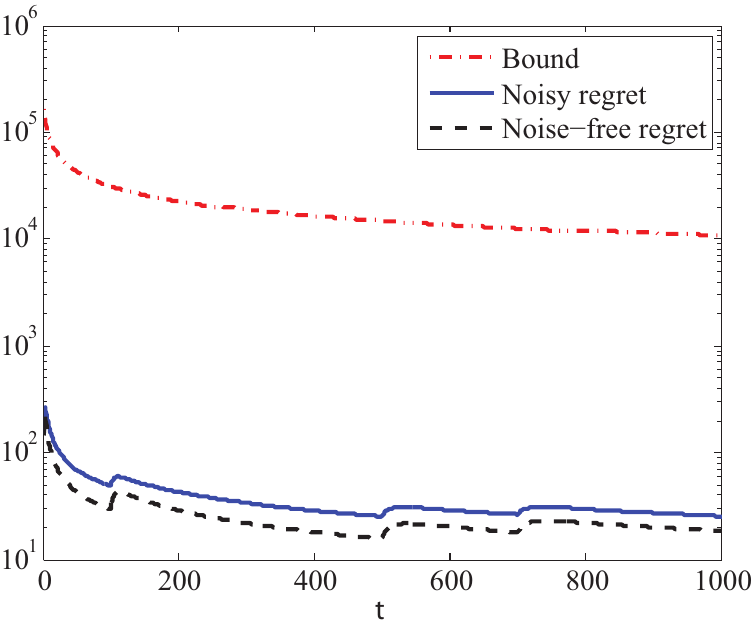} \\
	(c) & (d) & (e)		
\end{tabular}
\caption{\small (a) Ground truth and (b) Estimated $\wh{\mu}_t$. The
  $\mu$ values correspond to Bernoulli means, where lighter colors
  depict higher probabilities.  (c) Evolution of the log-loss from
  noiseless observations.  (d) Evolution of the log-loss from noisy
  observations.  The spikes at the jump times ($t = 100, 500,$ and
  $700$, indicated by red squares) correspond to model changes. Note that the variance of the
  log-loss is larger for the noisy case.  (e) Per-round regret
  compared to theoretical upper bound. Again, the regret is larger for
  the noisy case.}
\label{fig:exp1_mu}
\end{figure*}

\noindent \textbf{Experiment 2a.}  We now consider the detection of
anomalies using Algorithm \ref{alg:label_eff}.
For this experiment, data is generated using the same model described in Experiment 1 above. Recall that the goal is to detect anomalies corresponding to observations which would be difficult to predict given past data. In this spirit, we define the 25 observations following each changepoint in $\mu^*_t$ to be ``true'' anomalies, and our goal is to detect them with minimal probability of error. The basic idea is that after this window of 25 time steps, observations corresponding to the new generative model should be predictable based on past data and no longer anomalous.

In this experiment, the
Forecaster queries the end-user for feedback when $| \zeta(\wh{p}_t) - \tau_t|$ is small.  As discussed in
Remark~\ref{rem:comp_issues}, we use the transformation $\zeta(s) = Cs$ with $C = \exp(220)$. 
Figure \ref{fig:exp2_alg6_7}(a) shows the result of this experiment. Here the declared anomalies are shown as black dots superimposed on
the log-loss, and the feedback times are depicted underneath. This result is described in more detail and compared with the result of Experiment 2b below.

\noindent \textbf{Experiment 2b.}  This experiment is run on exactly the same data as in Experiment 2a; the only difference is the feedback mechanism. In particular, we test Algorithm \ref{alg:anomaly_arbitrary}, which is designed for
the case when feedback is provided at arbitrary times.  In this simulation feedback is always provided when the Forecaster declares an event
anomalous and with $20\%$ probability if it misses an
anomaly.

Figure \ref{fig:exp2_alg6_7} shows the results of both experiments. For
both cases note that after the first jump (when feedback is first
received) the Forecaster adapts its threshold, allowing it to
dramatically increase its detection rate in subsequent jumps.
Specifically, Table~\ref{tab:errors} shows the number of detection
misses and false alarms for both Algorithms \ref{alg:label_eff} and
\ref{alg:anomaly_arbitrary}.  For comparison, we also show the same
performance measures for the best static threshold chosen in
hindsight with full knowledge of the anomalies (i.e.\ $\{y_t\}$).  In both experiments, \FHTAGN \  outperforms
static thresholding.  The number of false alarms is significantly
lower than that of detection misses due to the high initial value of
the threshold $\tau_t$, which is driven lower as feedback arrives.

\begin{figure*}[t]
\centering
\begin{tabular}{cc}  
\footnotesize{Label--efficient \FHTAGN} & \footnotesize{Arbitrarily spaced feedback \FHTAGN}\\
\includegraphics[height=2.2in]{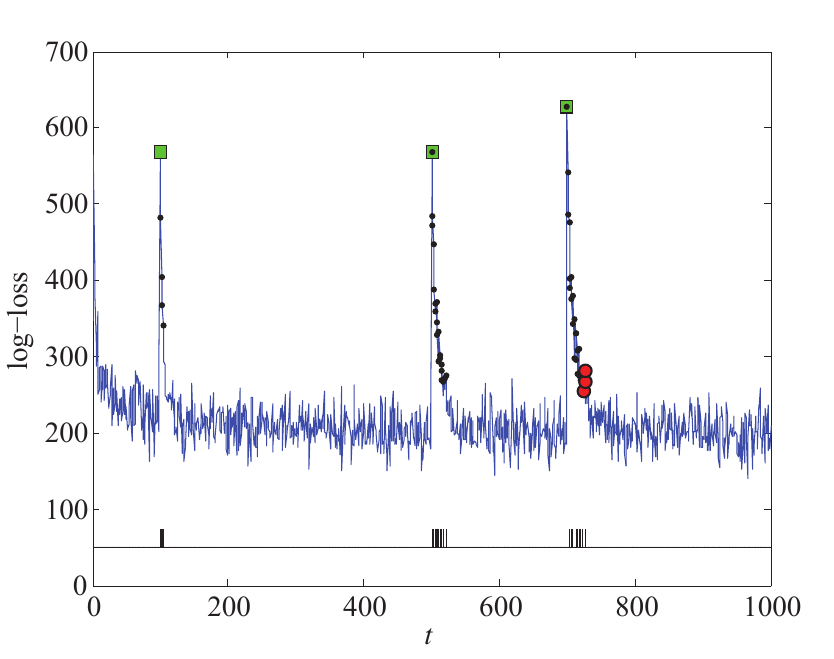}
&
\includegraphics[height=2.2in]{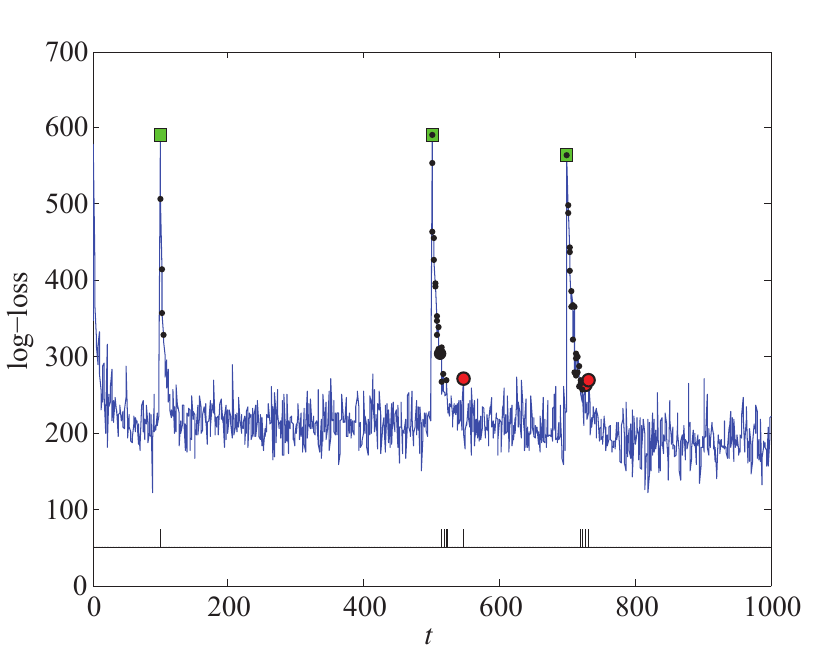} \\
(a) & (b)
\end{tabular}
\caption{\small (a) Anomalies detected (shown as small black dots) and false
  alarms (large red dots) by Algorithm \ref{alg:label_eff}, superimposed on the log--loss. Forecaster query times are shown below the log--loss. Jump times ($t = 100, 500,$ and $700$) are indicated by large green squares. (b) A similar plot of anomalies detected by Algorithm \ref{alg:anomaly_arbitrary}, with arbitrarily spaced feedback. In both cases, there were 25 true anomalies immediately following each jump. After the first jump, the Forecaster adapted its threshold, enabling it to dramatically increase its detection rate in subsequent jumps.}
\label{fig:exp2_alg6_7}
\end{figure*}

\begin{table}[htb]
\centering
   \begin{tabular}{|l|cc|cc|}
\hline
& \multicolumn{2}{|c|}{\em Label-efficient} &
\multicolumn{2}{|c|}{\em Arbitrary feedback times} \\ 
      & 	&	Best static & & Best static\\
      & \FHTAGN & threshold & \FHTAGN & threshold\\
\hline \hline
	Total Errors	& 30 & 44 & 34 & 46\\
	False alarms	& 3	& 8 & 3	& 9 \\
	Detection misses & 27	& 36 & 31	& 37\\
\hline
   \end{tabular} 
 \caption{ \label{tab:errors} Performance comparison of \FHTAGN\
     with anomaly detection using the best static threshold for
     Experiments 2a and 2b. \FHTAGN \ commits significantly fewer
     errors. For the Enron experiment, feedback was requested for 91 of the 902
     days considered, and only 523 of the 902 days had their text
     parsed. } 
 \end{table}

\textbf{Experiment 3.} Algorithm \ref{alg:label_eff} was applied to
the Enron email database \cite{Enron:04}, which consists of
approximately 500,000 e-mails involving 151 known employees and
more than 75,000 distinct addresses between the years 1998 and
2002. We use email timestamps in order to record users that were
active in each day, either sending or receiving emails. This was done
for 1,177 days, starting from Jan. 1, 1999. We removed days during
which no email correspondence occurred, and we consolidated each
weekend's emails into the preceding Friday's observation vector,
resulting in a total of 902 days in our dataset.  For each day $t$,
$x_t \in \{0,1\}^n$ is a binary vector indicating who sent or received
email that day.  In this setting, we let $\phi(x) = x$ leading to a
dimensionality of $n = d = 75511$. (There is no noise in this case,
since we can accurately identify sender and recipient email
addresses.)

Algorithms~\ref{alg:OCP_with_noise} and~\ref{alg:label_eff} were applied
to this dataset. The goal was to causally determine an accurate predictive model of email sender and recipient co-occurrences, and to identify anomalous periods of email activity using feedback.
For the hedging component, we use $\zeta(s) = C\log{s}$
with $C = 0.0079$, $\eta = 1450$, and $\tau_1=\wh{p}_1$.  Feedback was received when
requested according to Algorithm \ref{alg:label_eff}.  

The central idea here is that anomalous email discussion topics correspond to anomalous email sender and recipient co-occurrences. In this spirit, we generate
oracle or expert feedback (i.e., the $\{y_t\}$) based on the email {\em text} from the
difference in word counts between day $t$ and each of the previous 10
days, and average the result. Upon a feedback request at time $t$, we generate word
count vectors $h_t$ using the 12,000 most frequently appearing words
(to avoid memory issues and misspelled words) for days $t-10, \ldots, t-1, t$. Specifically, the mean wordcount deviation $e_t$ is computed as
$$
e_t=\frac{1}{10}\sum_{i =
    t-10}^{t-1}\|h_t-h_i\|_1
    $$
    where $\| \cdot \|_1$
is the $\ell_1$ norm. This can be considered a crude measure of temporal variation in text documents. When the deviation $e_t$ is
sufficiently high, we consider day $t$ to be anomalous according to
our expert system (i.e., $y_t = 1$). The deviation metric $e_t$ and the threshold determining $y_t$
are shown in the right upper plot in Figure \ref{fig:exp3_alg6}, but
note that only a fraction of these values need to be computed to run
\FHTAGN\ whenever feedback is requested.

\begin{figure*}[t]
\centering
\includegraphics[width=\textwidth]{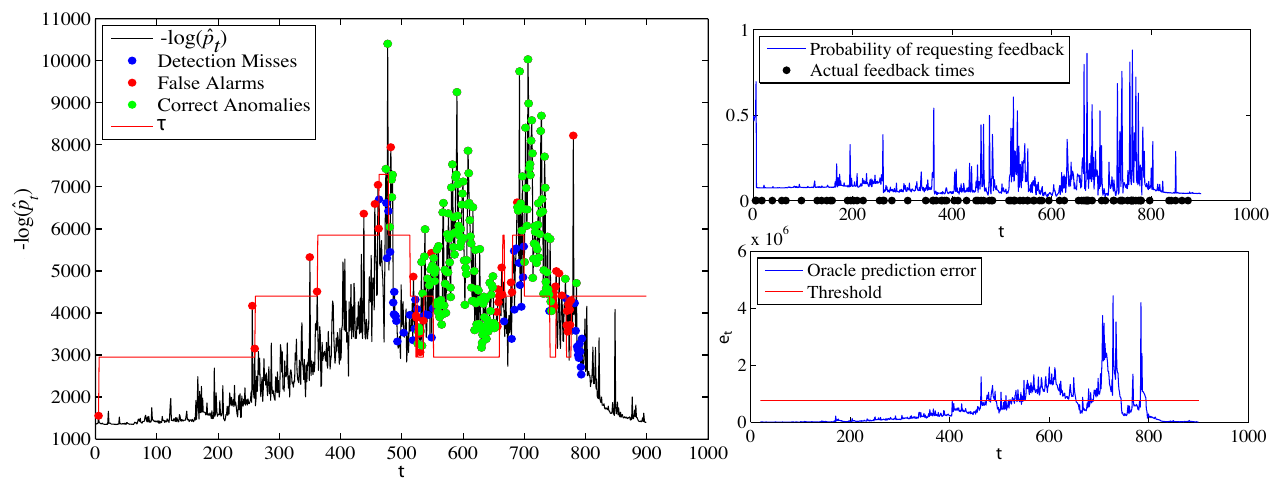}
\caption{\small Online anomaly detection results on Enron corpus. Left
  plot displays filtering output, locations of missed anomalies (as
  declared by our oracle), false positives, and correctly identified
  anomalies, as well as time-varying threshold $\tau$. Upper right
  plot displays the probability of requesting feedback where black
  circles indicate the locations where feedback was provided. Lower
  right plot displays oracle prediction error $e_t$ (from contextual
  evidence within a sliding window) compared to a static threshold to
  assign $y_t$.} 
\label{fig:exp3_alg6}
\end{figure*}
 
The results of Algorithm \ref{alg:label_eff} are summarized in Table
\ref{tbl:results} and Figure \ref{fig:exp3_alg6}. \FHTAGN\ performs
very well (in terms of detecting anomalies corresponding to the expert system designation with low probability of error) relative to a comparator online anomaly detection method
which consists of comparing $\wh{p}_t$ to the best static threshold,
chosen in hindsight with full knowledge of all filtering
outputs and feedback.
The left plot in Figure \ref{fig:exp3_alg6} shows the time varying
threshold $\tau_t$ in response to user feedback. In this experiment,
ground truth anomalies do not always correspond to large values of
$-\log{\wh{p}_t}$ but rather to the degree to which contextual evidence
differs from recent history. With a $10$-day memory, the
notion of what constitutes an anomaly is constantly evolving, and
$\tau_t$ adjusts to reflect the pattern.  The lower right plot
describes the probability of requesting feedback over time with the
days on which feedback was requested indicated with black dots. This
plot suggests that feedback is more likely to be requested on days
where $\zeta(\wh{p}_t)$ and $\tau_t$ have similar magnitude, which is
expected. Feedback was requested $91$ out of $902$ days, and because of
the sliding window used by our oracle to determine the true labels
$y_t$, a total of $523$ of the $902$ days required text parsing (and,
generally speaking, any overhead associated with decrypting,
transcribing, or translating documents).

\begin{table}[htb]
   \centering
   \begin{tabular}{|lcc|}
\hline
      & 	&	Best static\\
      & \FHTAGN & threshold \\
\hline \hline
	Total Errors	& 73 & 143 \\
	False alarms	& 35	& 96 \\
	Detection misses & 38	& 47 \\
\hline
   \end{tabular}
   \caption{\label{tbl:results} Performance comparison for \FHTAGN\ and
     the best static threshold on Enron data set. Feedback was requested for 91 of the 902
     days considered, and only 523 of the 902 days had their text
     parsed.}
\end{table}

Some of the most anomalous events detected by our proposed approach
correspond to historical events, as summarized in Table~\ref{tbl:enron}. These examples indicate that the anomalies in social network
communications detected by \FHTAGN\ are indicative of anomalous events
of interest to the social network members.
\begin{table}[htb]
\centering
\begin{tabular}{|lp{2.5in}|}
\hline
Date & Significance \\ \hline \hline
Dec. 1, 2000 & Days before ``California faces unprecedented
  energy alert'' (Dec. 7) and energy commodity trading deregulated in
  Congress. (Dec. 15) \cite{pbs_timeline}.\\ \hline
May 9, 2001 & ``California Utility Says Prices of Gas Were
  Inflated'' by Enron collaborator El Paso \cite{nytimes_inflated},
  blackouts affect upwards of 167,000 Enron customers \cite{cnn}.\\ \hline
Oct. 18, 2001 & Enron reports \$618M third quarter loss,
  followed by later major correction \cite{ft}.\\
\hline
\end{tabular}
\caption{\label{tbl:enron}Significant dates in Enron's history and our
analysis.}
\end{table}

\section{Conclusion}

We have proposed and analyzed a methodology for sequential (or online)
anomaly detection from an individual sequence of potentially noisy
observations in the setting when the anomaly detection engine can
receive external feedback confirming or disputing the engine's
inference on whether or not the current observation is anomalous
relative to the past. Our methodology, dubbed \FHTAGN\ for Filtering
and Hedging for Time-varying Anomaly recoGNition, is based on the
filtering of noisy observations to estimate the belief about the next
clean observation, followed by a threshold test. The threshold is
dynamically adjusted, whenever feedback is received and the engine has
made an error, which constitutes the hedging step. Our analysis of the
performance of \FHTAGN\ was carried out in the individual sequence
framework, where no assumptions were made on the mechanism underlying
the evolving observations. Thus, performance was measured in terms of
{\em regret} against the best {\em offline} (nonsequential) method for
assigning beliefs to the entire sequence of {\em clean} observations
and then using these beliefs and the feedback (whenever available) to
set the best critical threshold. The design and analysis of both
filtering and hedging was inspired by recent developments in online
convex programming.

One major drawback of the proposed filtering step is the need to
compute the log partition function. While closed-form expressions are
available for many frequently used models (such as Gaussian MRFs),
computing log partitions of general pairwise Markov random fields is
intractable \cite{WaiJor08}. While there exist a variety of techniques
for approximate computation of log partition functions, such as the
log determinant relaxation \cite{WaiJor06,BanGhaAsp08}, these
techniques themselves involve solving convex programs. This may not be
an issue in the offline (batch) setting; however, in a sequential
setting, computations may have to be performed in real
time. Therefore, an important direction for future research is to find
ways to avoid computing (or approximating) log partition functions in
the filtering step, perhaps by replacing the full likelihood with an
appropriate ``pseudo-likelihood'' \cite{RavWaiLaf09,HoeTib09}.

\appendix

\subsection{Proof of Theorem~\ref{thm:log_regret}}
\label{app:proof_log_regret}

For each $t$, let us use the shorthand $\lf_t(\theta)$ to denote the
filtering loss $\lf(\theta,z_t)$, $\theta \in \Lambda$. We start by
observing that, for any $\theta,\theta' \in \Theta$ we
have\footnote{In the terminology of \cite{BarHazRak08},
  (\ref{eq:strong_convexity}) means that the function $\theta \mapsto
  \lf_t(\theta)$ is {\em strongly convex} w.r.t.\ the Bregman
  divergence $D_\Phi(\theta,\theta') \equiv D(\theta' \| \theta)$ with
  constant $1$. In fact, their condition for strong convexity holds
  here with equality.}
\begin{align}
& \lf_t(\theta) - \left[ \lf_t(\theta') + \Ave{ \nabla \lf_t(\theta'), \theta - \theta' } \right]
\nonumber \\
& = -\ave{ \theta, h(z_t) } + \Phi(\theta) - \big[ - \ave{ \theta', h(z_t) } + \Phi(\theta') \nonumber\\
& \qquad \qquad - \ave{ h(z_t), \theta - \theta' } + \ave{ \nabla\Phi(\theta'), \theta - \theta' } \big] \nonumber \\
& = \underbrace{\ave{ \theta' - \theta, h(z_t) } + \ave{ \theta - \theta', h(z_t) }}_{=0} \nonumber\\
& \qquad \qquad + \Phi(\theta) - \Phi(\theta') - \ave{ \nabla \Phi(\theta'), \theta - \theta' } \nonumber \\
& \equiv D(\theta' \| \theta). \label{eq:strong_convexity}
\end{align}
In particular, using \eqref{eq:strong_convexity} with $\theta' =
\wh{\theta}_t$, we can write
\begin{align}\label{eq:1st_order_sc}
\lf_t(\wh{\theta}_t) - \lf_t(\theta) = - \ave{\nabla \lf_t(\wh{\theta}_t),\theta - \theta_t} - D(\wh{\theta}_t \| \theta).
\end{align}
Now, by our hypothesis on $\Lambda$, the Legendre potential $\Phi$ is
strongly convex w.r.t.\ the Euclidean norm $\| \cdot \|$ with constant
$\alpha = 2H$. Indeed, for any $\theta,\theta' \in \Lambda$ we have
\begin{align*}
&\Phi(\theta) - \Phi(\theta') - \ave{ \nabla \Phi(\theta'), \theta - \theta' } \nonumber\\
&\qquad = \frac{1}{2} \ave{ \theta - \theta', \nabla^2 \Phi(\theta'') (\theta - \theta') } \nonumber\\
&\qquad \ge H \| \theta - \theta' \|^2,
\end{align*}
where in the second step $\theta''$ is some point on the line segment joining $\theta$ and $\theta'$, and the last step follows from the fact that $\nabla^2 \Phi(\theta'') \succeq 2HI_{d \times d}$ for any $\theta'' \in \Lambda \subset \Theta_H$. Thus, we can apply Lemma~\ref{lm:basic_regret_bound} to \eqref{eq:1st_order_sc} to get
\begin{align}
& \lf_t(\wh{\theta}_t) - \lf_t(\theta) \nonumber\\
& = - \ave{\nabla \lf_t(\wh{\theta}_t),\theta - \theta_t} - D(\wh{\theta}_t \| \theta)\nonumber\\
&\le \frac{1}{\eta_t} \left( D(\wh{\theta}_t \| \theta) - D(\wh{\theta}_{t+1} \| \theta) \right)  + \frac{\eta_t}{4H} \| \nabla \lf_t(\wh{\theta}_t) \|^2 - D(\wh{\theta}_t \| \theta).\label{eq:dlf}
\end{align}
Now, if we define
\begin{align*}
\Delta_t \deq \begin{cases}
0, & t = 1 \\
\frac{1}{\eta_{t-1}}D(\wh{\theta}_t\|\theta), & t \ge 2
\end{cases}
\end{align*}
then we can rewrite \eqref{eq:dlf} as
\begin{align*}
 \lf_t(\wh{\theta}_t) - \lf_t(\theta) &\le \Delta_t - \Delta_{t+1} + \frac{\eta_t}{4H} \| \nabla \lf_t(\wh{\theta}_t) \|^2 \nonumber \\
& \qquad \qquad - D(\wh{\theta}_t \| \theta) + \frac{1}{\eta_t} D(\wh{\theta}_t \|\theta) - \Delta_t \nonumber\\
& = \Delta_t - \Delta_{t+1} + \frac{\eta_t}{4H} \| \nabla \lf_t(\wh{\theta}_t) \|^2,
\end{align*}
where in the last step we have used the fact that, with $\eta_t = 1/t$, $\frac{1}{\eta_t}D(\wh{\theta}_t \| \theta) - \Delta_t = D(\wh{\theta}_t \| \theta)$ for all $t$. Moreover, because
$$
\| \nabla \lf_t(\wh{\theta}_t) \| \le \| h(z_t) \| +  \| \nabla \Phi(\wh{\theta}_t) \| \le 2(K(z^T) + M),
$$
we get
\begin{align*}
\lf_t(\wh{\theta}_t) - \lf_t(\theta) \le \Delta_t - \Delta_{t+1} + \frac{(K(z^T) + M)^2 \eta_t}{H}.
\end{align*}
Summing from $t=1$ to $t=T$, we obtain
\begin{align*}
& \sum^T_{t=1} \lf(\wh{\theta}_t,z_t) - \sum^T_{t=1} \lf(\theta,z_t) \\
&\qquad\le \sum^T_{t=1} \left(\Delta_t - \Delta_{t+1}\right) + \frac{(K(z^T)+M)^2}{H} \sum^T_{t=1} \eta_t \\
&\qquad= \Delta_1 - \Delta_{T+1} + \frac{K+L)^2}{H}\sum^T_{t=1} \frac{1}{t} \\
&\qquad\le  \frac{(K(z^T)+M)^2}{H} \log(T+1),
\end{align*}
where in the last line we have used the estimate $\sum^T_{t=1} t^{-1} \le 1 + \int^T_1 t^{-1} dt = \log T+1$.

\subsection{Proof of Theorem~\ref{thm:root_regret}}
\label{app:proof_root_regret}

The main idea of the proof is similar to that in Theorem~\ref{thm:log_regret}, except that now care must be taken in dealing with the time-varying comparison strategy $\bd{\theta} = \{\theta_t\}$. Using \eqref{eq:dlf} with $\theta = \theta_t$ and the fact that $D(\cdot \| \cdot) \ge 0$, we can write
\begin{align}
\lf_t(\wh{\theta}_t) - \lf_t(\theta_t) &\le \frac{1}{\eta_t}\left(D(\wh{\theta}_t \| \theta_t) - D(\wh{\theta}_{t+1} \| \theta_t) \right)\nonumber \\
&\quad \quad + \frac{\eta_t}{4H} \| \nabla \lf_t(\wh{\theta}_t) \|^2 - D(\wh{\theta}_t \| \theta_t). \label{eq:time_varying_dlf}
\end{align}
Let us define
\begin{align*}
\Delta'_t \deq \begin{cases}
0, & t = 1 \\
\frac{1}{\eta_{t-1}} D(\wh{\theta}_t \| \theta_t), & t \ge 2
\end{cases}
\end{align*}
and $\Gamma_t \deq D(\wh{\theta}_{t+1} \| \theta_{t+1}) - D(\wh{\theta}_{t+1} \| \theta_t)$. Then we can rewrite \eqref{eq:time_varying_dlf} as
\begin{align*}
 \lf_t(\wh{\theta}_t) - \lf_t(\theta_t)  &\le \Delta'_t - \Delta'_{t+1} + \frac{1}{\eta_t} \Gamma_t + \frac{\eta_t}{4H} \| \nabla \lf_t(\wh{\theta}_t) \|^2 \\
& \qquad \qquad - D(\wh{\theta}_t \| \theta_t) + \frac{1}{\eta_t} D(\wh{\theta}_t \| \theta_t) - \Delta'_t \\
& \le \Delta'_t - \Delta'_{t+1} + \frac{1}{\eta_t}\Gamma_t + \frac{\eta_t}{4H} \| \nabla \lf_t(\wh{\theta}_t) \|^2,
\end{align*}
where the last step uses the easily checked fact that, with the choice $\eta_t = 1/\sqrt{t}$, 
\begin{align*}
\frac{1}{\eta_t} D(\wh{\theta}_t \| \theta_t) -\Delta'_t = (\sqrt{t} - \sqrt{t-1})D(\wh{\theta}_t \| \theta_t) \le D(\wh{\theta}_t \| \theta_t).
\end{align*}
Next, we have
\begin{align*}
\Gamma_t &= \Phi(\theta_{t+1}) - \Phi(\wh{\theta}_{t+1}) - \Ave{ \nabla \Phi(\wh{\theta}_{t+1}), \theta_{t+1} - \wh{\theta}_{t+1} } \\
& \qquad - \Phi(\theta_t) + \Phi(\wh{\theta}_{t+1}) + \Ave{ \nabla \Phi(\wh{\theta}_{t+1}), \theta_t - \wh{\theta}_{t+1} } \\
&= \Phi(\theta_{t+1}) - \Phi(\theta_t) + \Ave{ \nabla \Phi(\wh{\theta}_{t+1}), \theta_t - \theta_{t+1} } \\
&\le 4 M \left\| \theta_t - \theta_{t+1} \right\|.
\end{align*}
Moreover, just as in the proof of Theorem~\ref{thm:log_regret}, we have
$$
\| \nabla \lf_t(\wh{\theta}_{t})  \|^2 \le 4(K(z^T)+M)^2.
$$
Combining everything and summing from $t=1$ to $t=T$, we obtain
\begin{align*}
& \sum^T_{t=1} \lf(\wh{\theta}_t,z_t) - \sum^T_{t=1} \lf(\theta_t,z_t) \\
&\qquad \le \sum^T_{t=1} \left(\Delta'_t - \Delta'_{t+1}\right) +4 M\sum^T_{t=1} \frac{1}{\eta_t} \| \theta_t - \theta_{t+1} \| \\
& \qquad \qquad + \frac{(K(z^T)+M)^2}{H}\sum^T_{t=1} \eta_t \\
&\qquad \le \frac{4 M}{\eta_T} V_T(\bd{\theta}) + \frac{(K(z^T)+M)^2}{H}\sum^T_{t=1} \eta_t \\
&\qquad \le 4 M \sqrt{T} V_T(\bd{\theta}) + \frac{(K(z^T)+M)^2}{H} (2\sqrt{T}-1).
\end{align*}
In the last line, we have used the estimate $\sum^T_{t=1} t^{-1/2} \le 1 + \int^T_1 t^{-1/2}dt = 2\sqrt{T} - 1$.

\subsection{Proof of Lemma~\ref{lem:martingale}}
\label{app:proof_martingale}

For each $t$, we have
\begin{align*}
&\expect\left[\Df_{\btheta,t+1}(x^{t+1},z^{t+1})|\cR_t\right] \\
&\quad= \expect \Bigg[ \sum_{s=1}^{t+1} \ave{
  \theta_s,h(z_s)-\phi(x_s) } \Bigg| \cR_t \Bigg]\\
&\quad= \expect \left[ \ave{
  \theta_{t+1},h(z_{t+1})-\phi(x_{t+1}) } | \cR_t \right] \\
&\quad \quad \quad \quad \quad \quad + \expect \Bigg[ \sum_{s=1}^{t} \ave{
  \theta_s,h(z_s)-\phi(x_s) } \Bigg| \cR_t \Bigg]\\
&\quad= \ave{ \theta_{t+1}, \expect [h(z_{t+1})|\cR_t] - \phi(x_{t+1}) } \\
&\quad \quad \quad \quad \quad \quad +  \sum_{s=1}^{t} \ave{
  \theta_s,h(z_s)-\phi(x_s) }  \\
&\quad= 0 + \Df_{\btheta,t}(x^t,z^t)
\end{align*}
where in the third step we used the fact that $\theta_{t+1}$, $\{\theta_s\}_{s \le t}$, and $\{z_s\}_{s \le t}$ are $\cR_t$-measurable, and in the last step we used the fact that $\expect[ h(z_{t+1}) | \cR_t] = \phi(x_{t+1})$. Thus, $\left\{\Df_{\btheta,t}(x^t,z^t), \cR_t\right\}_{t\geq0}$, with $\Df_{\btheta,0}(x^0,z^0) \equiv 0$ and $\cR_0$ the trivial $\sigma$-algebra, is a zero-mean martingale, and the desired result follows.

\subsection{Proof of Theorem~\ref{thm:exact_regret}}
\label{app:proof_exact_hedge}

We closely follow the proof of Theorem~12.1 in \cite{CesLug06}, except that we use Lemma~\ref{lm:basic_regret_bound} to highlight the role of mirror descent and to streamline the argument. Let
$\ell'_t(\tau)$ denote the subgradient of $\tau \mapsto
\ell_t(\tau)$. Note that when $\ell_t(\tau) > 0$, $\ell'_t(\tau) =
-y_t$. Thus, when $\wh{y}_t \neq y_t$, the Forecaster implements the
projected subgradient update $\tau_{t+1} = \Pi\big(\tau_t - \eta
\ell'_t(\tau_t)\big)$. Thus, whenever $\wh{y}_t
\neq y_t$, we may use Lemma~\ref{lm:basic_regret_bound}:
\begin{align}
  \ell'_t(\tau)(\tau_t - \tau) & = (\tau - \tau_t)y_t \nonumber \\
& \le \frac{1}{2\eta} \Big((\tau - \tau_t)^2 - (\tau - \tau_{t+1})^2  \Big) + \frac{\eta}{2} |\ell'_t(\tau_t)|^2 \nonumber\\
& = \frac{1}{2\eta} \Big((\tau - \tau_t)^2 - (\tau - \tau_{t+1})^2  \Big) + \frac{\eta}{2}, \label{eq:step4}
\end{align}
where the inequalities hold for every $\tau$. Now, at any step at which $\wh{y}_t \neq y_t$, i.e., $\sgn(\tau_t - \zeta_t) \neq y_t$, the hinge loss $\ell_t(\tau) = (1 - (\tau - \zeta_t)y_t)_+$ obeys the bound
\begin{align}
1 - \ell_t(\tau) &= 1 - (1 - (\tau - \zeta_t)y_t)_+ \nonumber\\
&\le (\tau - \zeta_t)y_t \nonumber\\
&= -(\tau  - \zeta_t) \ell'_t(\tau_t).
\end{align}
Therefore, when $\wh{y}_t \neq y_t$, we have
\begin{align}
1 - \ell_t(\tau) &\le
 (\tau - \tau_t) y_t + \underbrace{(\tau_t - \zeta_t) y_t}_{< 0}
\nonumber\\
&\le \frac{1}{2\eta}  \left[(\tau - \tau_t)^2 - (\tau - \tau_{t+1})^2  \right] + \frac{\eta}{2},
\label{eq:hinge_bound_1}
\end{align}
where we used the fact that $\wh{y}_t \neq y_t$ implies that
$\sgn(\tau_t - \zeta_t) \neq y_t$, so that $(\tau_t - \zeta_t) y_t < 0$. Note
also that when $\wh{y}_t = y_t$, we will have $\td{\tau}_{t+1} =
\tau_t$, and since $\tau_t \in [\tau_{\min},\tau_{\max}]$, $\tau_{t+1}
= \Pi(\td{\tau}_{t+1}) = \tau_t$. Thus, the very last expression in
(\ref{eq:hinge_bound_1}) is identically zero when $\wh{y}_t =
y_t$. Hence, we get the bound
\begin{align*}
(1-\ell_t(\tau))1_{\{ \wh{y}_t \neq y_t\}} \le \frac{1}{2\eta} \left[(\tau - \tau_t)^2 - (\tau - \tau_{t+1})^2 \right] + \frac{\eta}{2}
\end{align*}
that holds for all $t$. Summing from $t=1$ to $t=T$ and rearranging, we get
\begin{align*}
\sum^T_{t=1} 1_{\{\wh{y}_t \neq y_t\}}
& \le \sum^T_{t=1} \ell_t(\tau) + \frac{1}{2\eta} (\tau - \tau_1)^2 + \frac{T\eta}{2} \\
& \le  \sum^T_{t=1} \ell_t(\tau) + \frac{(\tau_{\max} - \tau_{\min})^2}{2\eta} + \frac{T\eta}{2}.
\end{align*}
Choosing $\eta = (\tau_{\max} - \tau_{\min})/\sqrt{T}$, we obtain the regret bound (\ref{eq:exact_regret}).

\subsection{Proof of Theorem~\ref{thm:label_eff}} 
\label{app:proof_label_eff}

We closely follow the proof of Theorem~12.5 in \cite{CesLug06}, but, again, we use Lemma~\ref{lm:basic_regret_bound} to streamline and simplify the argument. Introduce the random variables $M_t = 1_{\{\wh{y}_t \neq y_t\}}$. Then, whenever $M_t = 1$, we have $1-\ell_t(\tau) \le (\tau - \zeta_t)y_t$.
When $M_tU_t = 1$ (i.e.,~when $\tau_t$ is updated to $\tau_{t+1}$), we can use Lemma~\ref{lm:basic_regret_bound} and obtain
\begin{align*}
1-\ell_t(\tau) &\le  (\tau_t - \zeta_t) y_t +  (\tau - \tau_t) y_t \\
&\le (\tau_t - \zeta_t)y_t + \frac{1}{2\eta}\left[(\tau - \tau_t)^2 - (\tau - \tau_{t+1})^2\right] + \frac{\eta}{2}.
\end{align*}
From this, we obtain the inequality
\begin{align*}
& (1 + |\zeta_t - \tau_t|) M_t U_t \\
&\qquad \le \ell_t(\tau) + \frac{1}{2\eta}\left[(\tau - \tau_t)^2 - (\tau - \tau_{t+1})^2 \right] + \frac{\eta}{2},
\end{align*}
which holds for all $t$. Indeed, if $M_t U_t = 0$, the left-hand side
is zero, while the right-hand side is greater than zero since
$\ell_t(\tau) \ge 0$ and $\tau_t = \tau_{t+1}$. If
$M_t U_t = 1$, then $y_t (\tau_t - \zeta_t) = - (\tau_t - \zeta_t)
\sgn(\tau_t - \zeta_t) = - |\zeta_t - \tau_t|$. Summing over $t$, we
get
\begin{align*}
& \sum^T_{t=1} (1+|\zeta_t - \tau_t|) M_t U_t \le \sum^T_{t=1} \ell_t(\tau) \\
 & \qquad + \frac{1}{2\eta} \sum^T_{t=1} \left[(\tau - \tau_t)^2 - (\tau - \tau_{t+1})^2 \right] + \frac{T\eta}{2}.
\end{align*}
We now take expectation of both sides. Let ${\cal U}_t$ denote the $\sigma$-algebra generated by $U_1,\ldots,U_t$, and let $\expect_t[\cdot]$ denote the conditional expectation $\expect[\cdot|{\cal U}_{t-1}]$. Note that $M_t$ and $|\zeta_t - \tau_t|$ are measurable w.r.t.\ $\cR_{t-1}$, since both of them depend on $U_1,\ldots,U_{t-1}$,  and that $\expect_t U_t = 1/(1+|\zeta_t - \tau_t|)$. Hence
\begin{align*}
\expect \left[ \sum^T_{t=1} (1+ |\zeta_t - \tau_t|) M_t U_t \right] &=
\expect \left[\sum^T_{t=1} (1+|\zeta_t - \tau_t|) M_t \expect_t U_t
\right] \\
&= \expect \left[\sum^T_{t=1} M_t \right].
\end{align*}
Using the same argument as before with $\eta = (\tau_{\max} - \tau_{\min})/\sqrt{T}$, we obtain
$$
\expect\left[\sum^T_{t=1} 1_{\{\wh{y}_t \neq Y_t \}} \right] \le \sum^T_{t=1} \ell_t(\tau) + (\tau_{\max} - \tau_{\min})\sqrt{T},
$$
and the theorem is proved.

\section*{Acknowledgments}

\noindent The authors would like to thank Sasha Rakhlin for helpful discussions, and two anonymous referees for their suggestions and comments that helped improve the presentation.

\bibliography{anomaliesFiltering.bbl}

\begin{IEEEbiographynophoto}{Maxim Raginsky} (S'99--M'00) received the B.S. and M.S. degrees in 2000 and the Ph.D. degree in 2002 from Northwestern University, Evanston, IL, all in electrical engineering. He has held research positions with Northwestern, the University of Illinois at Urbana-Champaign (where he was a Beckman Foundation Fellow from 2004 to 2007), and Duke University. In 2012, he has returned to UIUC, where he is currently an Assistant Professor with the Department of Electrical and Computer Engineering and the Coordinated Science Laboratory. His research interests lie at the intersection of information theory, machine learning, and control.\end{IEEEbiographynophoto}

\begin{IEEEbiographynophoto}{Rebecca M.~Willett} (S'01--M'05--SM'11) received the Ph.D. degree in
electrical and computer engineering from Rice University, Houston, TX,
in 2005. She is currently an Assistant Professor with the Department
of Electrical and Computer Engineering, Duke University, Durham, NC.
She has also held visiting researcher positions with the Institute for
Pure and Applied Mathematics, University of California, Los Angeles,
in 2004, the University of Wisconsin-Madison, from 2003 to 2005, the
French National Institute for Research in Computer Science and Control
(INRIA), Paris, France, in 2003, and the Applied Science Research and
Development Laboratory, GE Healthcare, in 2002. Her research interests
include network and imaging science with applications in medical
imaging, wireless sensor networks, astronomy, and social networks.
Prof. Willett is a member of the Defense Advanced Research Projects
Agency Computer Science Study Group. She was the recipient of the
National Science Foundation CAREER Award in 2007 and the Air Force
Office of Scientific Research Young Investigator Program Award in
2010.
\end{IEEEbiographynophoto}

\begin{IEEEbiographynophoto}{Corinne Horn} received the B.S.E. degree in electrical and computer
engineering from Duke University, Durham, NC, in 2011. She is
currently a graduate student in the Department of Electrical
Engineering, Stanford University, Stanford, CA.
\end{IEEEbiographynophoto}

\begin{IEEEbiographynophoto}{Jorge Silva} (M'00) received his EE, MSc and PhD in electrical and computer engineering from Instituto Superior T\'{e}cnico (IST), Lisbon, Portugal, in 1993, 1999 and 2007, respectively. He was a researcher at Instituto de Engenharia de Sistemas e Computadores (INESC) in 1993--1996, and at the Instituto de Sistemas e Rob\'{o}tica (ISR), Lisbon, in 2003--2007. He held teaching positions at Instituto Superior de Engenharia de Lisboa (ISEL) in 1996--2007. In the same period, he did consulting and R\&D work for major Portuguese utility and transportation companies. He is now a Research Scientist, Sr., at Duke University, where he is developing estimation methods for very high-dimensional spaces. His research interests include signal processing, manifold learning, computer vision and social network analysis.
\end{IEEEbiographynophoto}

\begin{IEEEbiographynophoto}{Roummel F.~Marcia} (M'08) received  the Ph.D. degree in mathematics from the University of California, San Diego, in 2002.  He is currently an Assistant Professor of applied mathematics with the School of Natural Sciences, University of California, Merced.  He was a Computation and Informatics in Biology and Medicine Postdoctoral Fellow with the Department of Biochemistry, University of Wisconsin-Madison, and a Research Scientist with the Department of Electrical and Computer Engineering, Duke University. His research interests include nonlinear optimization, numerical linear algebra, signal and image processing, and computational biology.\end{IEEEbiographynophoto}

\end{document}